\documentclass[lettersize,journal]{IEEEtran}
\usepackage{amsmath,amsfonts}
 \usepackage[ruled, vlined, linesnumbered]{algorithm2e}
\usepackage{array}
\usepackage[caption=false,font=normalsize,labelfont=sf,textfont=sf]{subfig}
\usepackage{textcomp}
\usepackage{stfloats}
\usepackage{url}
\usepackage{verbatim}
\usepackage{graphicx}
\usepackage{cite}
\usepackage{color}
\usepackage{colortbl}
\usepackage[colorlinks=true]{hyperref}
\usepackage{xcolor}
\usepackage{multirow}
\usepackage{booktabs}
\usepackage{bbding}
\usepackage{makecell}

\begin{document}

\title{Advancing Pre-trained Teacher: Towards Robust Feature Discrepancy for Anomaly Detection}

\author{IEEE Publication Technology,~\IEEEmembership{Staff,~IEEE,}

\author{Canhui Tang, 
       Sanping Zhou,~\IEEEmembership{Member,~IEEE},
       Yizhe Li, Yonghao Dong,
       Le Wang,~\IEEEmembership{Senior Member,~IEEE}
        % <-this % stops a space
\thanks{
This work was supported in part by National Science and Technology Major Project under Grant 2023ZD0121300, NSFC under Grants 62572384 and 12326608, Natural Science Foundation of Shaanxi Province under Grant 2022JC-41, Fundamental Research Funds for the Central Universities under Grant XTR042021005, and Guangdong Major Project of Basic and Applied Basic Research under Grant 2023B0303000009. ({\it Corresponding author: Sanping Zhou, E-mail: spzhou@mail.xjtu.edu.cn.})}
\thanks{
Canhui Tang, Sanping Zhou, Yizhe Li, Yonghao Dong, and Le Wang are all with the National Key Laboratory of Human-Machine Hybrid Augmented Intelligence, National Engineering Research Center for Visual Information and Applications, and Institute of Artificial Intelligence and Robotics, Xi'an Jiaotong University, Shaanxi 710049, China.}
}

}

% The paper headers
\markboth{IEEE TRANSACTIONS ON IMAGE PROCESSING}%
{Shell \MakeLowercase{\textit{et al.}}: A Sample Article Using IEEEtran.cls for IEEE Journals}

\maketitle

\begin{abstract}
With the wide application of knowledge distillation between an ImageNet pre-trained teacher model and a learnable student model,
unsupervised anomaly detection has
witnessed a significant achievement in the past few years. The success of this framework mainly relies on how to keep the feature discrepancy between the teacher and student model, in which it has two underlying sub-assumptions:~\textbf{(1)} {The teacher model can represent two separable distributions for the normal and abnormal patterns}, while~\textbf{(2)} the student model can only reconstruct the normal distribution. However, it still remains a challenging issue to maintain these ideal assumptions in practice. In this paper, we propose a simple yet effective two-stage industrial anomaly detection framework, termed AAND, which sequentially performs Anomaly Amplification and Normality Distillation to enhance the two assumptions.
In the first anomaly amplification stage, we propose a novel Residual Anomaly Amplification (RAA) module to advance the pre-trained teacher encoder with synthetic anomalies. It generates adaptive residuals to amplify anomalies while maintaining the feature integrity of pre-trained model. It mainly comprises a Matching-guided Residual Gate and an Attribute-scaling Residual Generator, which can determine the residuals' proportion and characteristic, respectively. In the second normality distillation stage, we further employ a reverse distillation paradigm to train a student decoder, in which a novel Hard Knowledge Distillation~(HKD) loss is built to better facilitate the reconstruction of normal patterns. Comprehensive experiments on the MvTecAD, VisA, and MvTec3D-RGB datasets show that our method achieves state-of-the-art performance. Our code is available at \href{https://github.com/Hui-design/AAND}{https://github.com/Hui-design/AAND}.
\end{abstract}


\section{Introduction}
% Industrial Anomaly Detection~(IAD) is a crucial task that focuses on detecting and localizing anomalies in images of industrial products. Due to the scarcity of anomaly samples, IAD is typically treated as an unsupervised problem, relying solely on normal samples for training. This setting requires IAD methods to capture and model the distribution of normal samples during training, enabling subsequent anomaly detection by identifying deviations from the learned distribution. In practice, feature extractors pre-trained on ImageNet~\cite{deng2009imagenet} have been widely applied in the IAD task~\cite{patchcore,RD,cflow}. The pre-trained model is mainly used to generate representations for industrial instances, allowing for the capture of diverse normal patterns and abnormal patterns to some extent. In particular, some approaches store representative features in memory banks~\cite{patchcore,M3DM}, and then treat the kNN distance as anomaly scores. Additionally, some methods employ additional decoders to learn the probability distribution using normalization flows~\cite{cflow,csflow}.

Industrial Anomaly Detection~(IAD) is a crucial task that focuses on detecting and localizing anomalies in images of industrial products. 
The scarcity of anomaly samples poses a challenge, leading IAD to be typically approached as an unsupervised problem that relies solely on normal samples for training. Moreover, industrial anomalies are hard to detect and localize due to their diverse and uncertain nature, which can range from subtle texture changes to significant structural defects.
In practice, feature extractors pre-trained on ImageNet~\cite{deng2009imagenet} have been widely applied in the IAD task~\cite{patchcore,RD,cflow}, in which they are utilized to generate representations for the diverse normal and abnormal patterns. By capturing and modeling the distribution of pre-trained features for normal samples, subsequent anomaly detection can be achieved by identifying deviations from the learned distribution.

\begin{figure}[t]
    \centering
\includegraphics[scale=0.68]{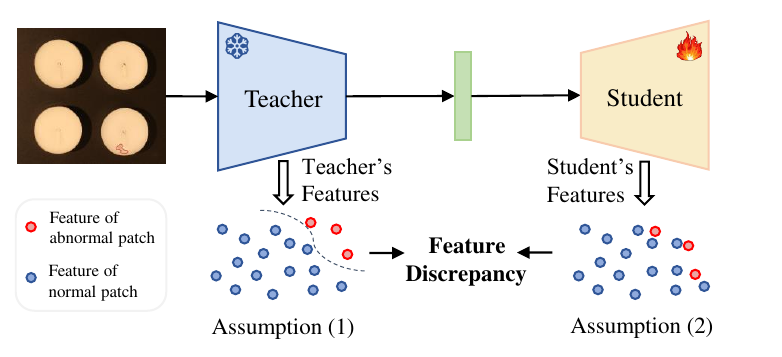}
    \caption{Illustration of Teacher-Student Feature Discrepancy, whose success relies on how to keep the feature discrepancy between teacher and student model. This line of methods has two underlying assumptions that have not been well addressed:~\textbf{(1)} {The teacher model can represent two separable distributions for the normal and abnormal patterns}; while~\textbf{(2)} the student decoder can only reconstruct the normal distribution.}
    \label{fig1}
\end{figure}

Recent advances~\cite{ST,MK,RD,RDplus,destseg} often adopt knowledge distillation to take advantage of the pre-trained model, which involves a multi-scale feature reconstruction of the pre-trained features. As shown in Fig.~\ref{fig1}, it solves the anomaly detection problem by focusing on the ``teacher-student discrepancy", \emph{i.e.}, the feature discrepancy between a fixed pre-trained teacher encoder and a learnable student decoder. To make the ``teacher-student discrepancy" hypothesis more robust, particularly ensuring larger discrepancies when anomalies are presented, two underlying sub-assumptions need to be satisfied:~\textbf{(1)} {The teacher model can represent two separable distributions for the normal and abnormal patterns}, while~\textbf{(2)} the student decoder can only reconstruct the normal distribution. 
% However, this hypothesis is hard to meet in practice, which requires researchers to improve it from different perspectives.
However, these ideal assumptions are hard to meet in practice, which requires researchers to improve them from different perspectives.

For the first assumption, it is important yet has been overlooked in current methods. As shown in Table~\ref{teacher_tab}, {we measure the cosine distance of multi-layer teacher features between normal patches and abnormal patches. We can observe a significant reduction in inter-class distance and a performance (I-AUC) drop} when the teacher model is randomly initialized in the well-known RD~\cite{RD} framework. Although the pre-trained model can alleviate this issue, it still remains a challenging issue to generate discriminative features in solving the IAD problem. To the best of our knowledge, the challenge mainly arises from two aspects:~\textbf{(1)} \emph{Difference between task objectives}. In pre-training on ImageNet, models aim to discriminate between different object categories; while in the IAD task, models need to discriminate between normal and abnormal instances within a single object category.~\textbf{(2)} \emph{Scarcity of abnormal data}. Abnormal objects are far less common than normal objects in the real world,  therefore the models have fewer opportunities to learn representations of anomalies during pre-training. Due to these factors, the pre-trained embeddings may become over-compressed and learn some irrelevant semantics to the IAD task, which limits the pre-trained teacher model's ability to discriminate between normal and abnormal industrial instances.
When it comes to the second assumption, earlier methods \cite{MK,RD} give a natural constraint, \emph{i.e.}, anomalies cannot reconstruct abnormal representations since they are absent in the training process.
To guarantee the student model consistently outputs normal patterns, some recent works~\cite{RDplus,destseg} impose stronger constraints on the student model to suppress anomaly signals. Despite these, less attention has been drawn to the reconstruction capacity of the student decoder in dealing with challenging normal patterns, such as the fine-grained normal textures
and rare normal patterns.

\begin{table}[t]
\setlength\tabcolsep{3.5pt}
\footnotesize
\centering
\caption{{Comparison results of teacher model on the VisA~\cite{visa} dataset. ``$\checkmark\kern-1.2ex\raisebox{1ex}{\rotatebox[origin=c]{125}{\textbf{--}}}$'' indicates the assumption is not fully guaranteed, and ``IC Distance'' represents the inter-class cosine distances of multi-layer features (1/2/3 layers). For fair comparison, the student model is set consistent.}}
\label{teacher_tab}
\begin{tabular}{cccc}
   \toprule
  \textbf{Teacher model} & \textbf{Assump.~(1)} & \textbf{IC Distance} & \textbf{I-AUC}~(\%) \\
  \midrule
  Random Teacher&$\times$&0.13 / 0.06 / 0.05&81.3\\
 Pre-trained Teacher~(RD)&$\checkmark\kern-1.2ex\raisebox{1ex}{\rotatebox[origin=c]{125}{\textbf{--}}}$&0.43 / 0.56 / 0.62&96.2\\
  Advanced Teacher (Ours) &\checkmark&\bf{0.52} / \bf{0.63} / \bf{0.67}&\bf{97.8}\\
   \bottomrule
\end{tabular}
\end{table}

To this end, we consider how to extend the teacher encoder’s discrimination capacity and how to enhance the student decoder's normality distillation ability. In this paper, we propose a two-stage framework to achieve robust feature discrepancy between the teacher and student model. Specifically, it consists of an anomaly amplification stage and a normality distillation stage, which address the first and second assumptions, respectively. In the first stage, we propose a novel Residual Anomaly Amplification~(RAA) module to advance the fixed teacher with the exposure of synthetic anomalies.  {Instead of proposing pioneering operators~\cite{yang2024kolmogorov,yang2024neural}, the primary novelty of our RAA lies in building a robust anomaly amplification mechanism that enhances the pre-trained teacher's discrimination ability using synthetic anomalies. Pre-trained features have learned a universal distribution of normal patterns, while over-reliance on synthetic anomalies may lead to 
catastrophic forgetting~\cite{forget} of this pre-trained knowledge.} To address 
this, our RAA module employs an adaptive residual learning mechanism. This module comprises two components: a matching-guided residual gate and an attribute-scaling residual generator, which determine the proportion and characteristics of residuals, respectively.  This model effectively amplifies the sensitivity to anomalies while maintaining the feature integrity of pre-trained model. In the second stage, we employ a reverse distillation paradigm to train a student decoder that can only reconstruct normal representations, in which a novel hard knowledge distillation~(HKD) loss is built to better facilitate the reconstruction of normal patterns. Without losing generality, the teacher-student discrepancy is utilized for anomaly detection during inference. Comprehensive experiments on the  MvTecAD, VisA, and MvTec3D-RGB datasets demonstrate the effectiveness of our method.

Our main contributions are as follows:
\begin{itemize}
\item We design a novel two-stage anomaly detection framework based on the two underlying assumptions of knowledge distillation, which can obtain robust feature discrepancy between the teacher and student model.
%\item We give an in-depth analysis of the implied two assumptions in distillation-based anomaly detection framework. Furthermore, we figure out that the first assumption is overlooked in previous work, which makes the teacher-student discrepancy unreliable.

%\item We design an anomaly amplification stage with a novel RAA module to extend the vanilla teacher network’s representation capacity. This stage helps amplify anomalies via residual generation while maintaining the integrity of the pre-trained model.

\item We design a novel residual anomaly amplification module in the anomaly amplification stage, which can enhance the discrimination capacity of the vanilla teacher encoder with the help of synthetic anomalies.
% via a residual learning mechanism.

\item We design a novel hard knowledge distillation loss in the normality distillation stage, which can enhance the reconstruction capability of the student decoder in dealing with challenging normal patterns.

%\item We propose a hard knowledge distillation loss that directs attention towards the challenging normal patterns, resulting in higher reconstruction accuracy for normal patterns without the need for additional blocks.
\end{itemize}

The rest of this paper is organized as follows. We review the related work in Section \ref{related}. Section \ref{method} describes the technical
details of our proposed AAND. Section \ref{exper} presents the
experiment details and results. Finally, we summarize the
paper in Section \ref{conclu}.

\section{Related Work}\label{related}
In this section, we review some related works on IAD. The current works can be categorized into four types: reconstruction-based, synthesis-based, embedding-based, and distillation-based methods.

\textbf{Reconstruction-based methods.} As anomaly detection is typically treated as an unsupervised problem, reconstruction-based methods have emerged as a natural approach. 
It aims to reconstruct normal inputs and assumes that the reconstruction error is larger for anomalies. Early methods often employ GAN~\cite{gan}, AE~\cite{AE,MemAE,AE-VAD,hou2021divide,AE-medical}, or VAE~\cite{vae,vae_video} to train reconstruction models. To make the anomaly pixels more difficult to reconstruct, RIAD~\cite{RIAD} proposes a reconstruction-by-inpainting approach, which removes partial pixels to reconstruct the images.
Inspired by the ability of latent diffusion model~\cite{latent,ho2020denoising} in generating high-quality
and diverse images, more and more recent approaches~\cite{wyatt2022anoddpm,scoreDDPM,removing,ddpm_recons} utilize diffusion models to model the image reconstruction as a denoising process. Nevertheless, the pixel-level reconstruction methods may easily generalize to anomalies, resulting in the well-reconstruction of some anomalies. To address this issue, some research has focused on exploring knowledge distillation, with a shift towards reconstructing inputs at the feature level.
 
% Instead of reconstructing low-level pixels, some methods~\cite{UniAD,FOD,fang2023fastrecon,ST, RD,RDplus} focus on reconstructing high-level pre-trained features, where knowledge distillation is a typical approach for feature reconstruction. This line of methods is distinguished for its robustness and effectiveness, and thus we adopt the knowledge distillation paradigm in our method.

\textbf{Synthesis-based methods.} Due to the lack of abnormal samples in training, it is challenging for the IAD task to find the boundary between normality and anomaly. Thus, synthesis-based methods propose to synthesize anomalies by introducing noises into normal images~\cite{zavrtanik2021draem,cutpaste,NSA,video_synthesis}. For example, in the case of DRAEM~\cite{zavrtanik2021draem}, textures are sampled from an external texture dataset~\cite{cimpoi2014describing}, and Perlin~\cite{perlin1985image} noise is employed to generate random anomaly masks for creating anomalies. Besides, Cutpaste~\cite{cutpaste} randomly selects augmented regions from an image and pastes it onto other random regions of the image. {NSA~\cite{NSA} innovatively employs Poisson image editing to seamlessly blend patches of varying sizes from different images, which provides an excellent anomaly synthesis method.}
Additionally, some methods synthesize noise at the feature level, such as SimpleNet~\cite{liu2023simplenet} and DSR~\cite{DSR}. 
% Synthesized anomalies are also used in embedding-based and distillation-based methods as auxiliary training. For example, RD++~\cite{RDplus} and DeSTSeg~\cite{destseg} utilize synthesis to encourage the network to reconstruct normal features from corrupted pseudo-abnormal regions. In addition, DeSTSeg~\cite{destseg} employs synthesis to train a segmentation network that effectively combines multi-level features. PNI~\cite{bae2023pni} employs synthesis to train a refinement network, which enhances the shape and edge details of the anomaly map. 
However, real-world anomalies exhibit a wide range of patterns and uncertainties that cannot be fully covered. Consequently, over-reliance on synthetic anomalies may overfit the seen anomalies, while getting bad performance on unseen anomalies. 
In our method, we utilize synthetic anomalies to amplify the sensitivity
of the pre-trained model to anomalies, while keeping its original discrimination ability on some unseen anomalies. 
To the best of our knowledge, we are the first to enhance the pre-trained teacher model with synthetic anomalies in solving the IAD problem.

% # Pre-trained models for AD
% # Anomaly Synthesis
% # Knowledge distillation based methods

% # Anomaly Detection
% # Knowledge distillation

\textbf{Embedding-based methods.}
Thanks to the development of models pre-trained on extensive external datasets, such {methods~\cite{wideresnet,zanella2024harnessing} }can effectively represent both normal and abnormal patterns for anomaly detection. In practice, the embedding-based methods usually utilize ImageNet pre-trained models to encode normal samples into high-dimensional feature spaces. For example, Padim~\cite{padim} calculates the inverse covariance matrix to model the normal distribution, Patchcore~\cite{patchcore} stores representative features in memory banks, and M3DM~\cite{M3DM} constructs multiple memory banks to store features from different modalities. What's different, some methods~\cite{differnet,cflow,csflow,BGAD} employ 
additional decoders to learn the distribution of normal samples using normalizing flows~\cite{nf}. For example, CS-Flow~\cite{csflow} proposes a simple framework that jointly processes multiple
feature maps of different scales. Besides, CFlow~\cite{cflow} models a conditional normalizing flow with a position encoding, and adopts a multi-scale generative decoder to estimate the likelihood of
encoded features. Since these methods largely rely on the pre-trained model, their performances are limited by the discrimination capacity of pre-trained model.

% The student model is trained to fit the normal data and its generated features are assumed to be discrepant for anomalies.
% \iffalse The training process involves a student model to reconstruct the multi-scale pre-trained embeddings of a teacher model on normal samples. \fi
\textbf{Distillation-based methods.}  This line of methods can combine the advantages of both reconstruction-based and embedding-based methods, which involve a student model to reconstruct the multi-scale pre-trained embeddings of a teacher model on normal samples. The mainstream methods can be roughly categorized into forward distillation and reverse distillation methods. In particular,
the forward distillation approaches~\cite{ST, MK} often adopt similar networks between student and teacher models to train the distillation process. Nevertheless, they suffer
from the similarity of student and teacher architecture, where the abnormal data is mapped closely, resulting in an undesirably small feature discrepancy for anomalies. To address this issue, RD~\cite{RD} adopts a reverse distillation paradigm. Rather than directly receiving raw images, the student network takes the one-class embedding of the teacher model as input, to restore the teacher's multi-scale representations. Besides, AST~\cite{AST} proposes a framework comprising a pre-trained encoder and two decoders, in which a normalizing flow decoder serves as a teacher for density estimation, while a conventional feed-forward network acts as a student. To guarantee the student network consistently outputs normal patterns, recent methods impose stronger constraints on the
student network. For example, RD++~\cite{RDplus} introduces multi-scale projection layers and incorporates several loss constraints, so as to facilitate the student model in suppressing abnormal signals from pseudo-abnormal regions. Besides, DeSTSeg~\cite{destseg} introduces a denoising stage to train a denoising student that reconstructs the normal patterns both for normal and abnormal inputs.

However, most distillation-based methods solely focus on the student model, while overlooking the limited discrimination ability of the teacher model. Unlike previous methods, we both focus on enhancing the teacher model's discrimination capacity and the student model's normality distillation ability.

\begin{figure*}[t]
    \centering
    \includegraphics[scale=1.0]{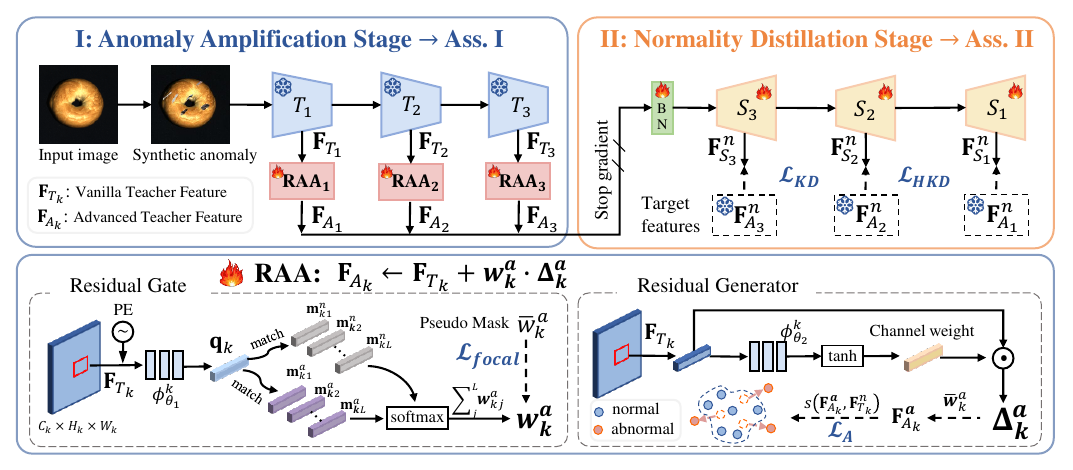}
    \caption{{Overview of our proposed AAND.~~\textbf{I.} Anomaly Amplification Stage for enhancing Assumption I (Ass. I): the vanilla teacher model encodes the input image into K-level features $\mathbf{F}_{T_k}$, and then these features are advanced to $\mathbf{F}_{A_k}$ through our proposed Residual Anomaly Amplification module (RAA). The RAA module generates adaptive residuals, effectively amplifying anomalies and preserving the integrity of the pre-trained model. It comprises a matching-guided residual gate and an attribute-scaling residual generator, which determine the proportion and characteristics of the residuals, respectively.~\textbf{II.} Normality Distillation Stage for enhancing Assumption II (Ass. II): a student model decodes features $\mathbf{F}_{S_k}$, and these features are trained to distill the representation of advanced teacher only on normal samples, where ``BN" denotes a bottleneck module and $\mathcal{L}_{HKD}$ represents our proposed Hard Knowledge Distillation loss. During inference, the ``teacher-student discrepancy'' is used for anomaly detection.} }
    \label{fig3}
\end{figure*}

\section{Method}\label{method}

% \textcolor{red}{
% \textbf{Preliminary. } In the context of unsupervised industrial anomaly detection, let $\mathcal{X}^{train} = \{\mathbf{I}_1^{n},\mathbf{I}_2^n ..., \mathbf{I}_{N}^{n}\}$ be a train set containing only normal images, and $\mathcal{X}^{test} = \{\mathbf{I}_1^{t},\mathbf{I}_2^{t} ..., \mathbf{I}_{M}^{t}\}$ be a test set including both normal and abnormal images. Given a test image $\mathbf{I^a}$ $\in \mathbb{R}^{C\times H\times W}$, The goal of anomaly detection is to predict an image-level score $\mathbf{S^a}$ $\in \mathbb{R}$ and a pixel-level score map $\mathbf{M^a}$ $\in \mathbb{R}^{H\times W}$, which identifies and localizes anomalies, respectively. }

% \subsection{Overview. } 
In the context of unsupervised industrial anomaly detection, let $\mathcal{X}^{train} = \{\mathbf{I}_1^{n},\mathbf{I}_2^n ..., \mathbf{I}_{N}^{n}\}$ be a training set containing \textbf{only normal images}, and $\mathcal{X}^{test} = \{\mathbf{I}_1^{t},\mathbf{I}_2^{t} ..., \mathbf{I}_{M}^{t}\}$ be a test set including both normal and abnormal images. The goal of anomaly detection is
to train a model to identify and localize anomalies in the test images.  

% As shown in Fig.~\ref{fig3}, we propose a simple yet effective two-stage anomaly detection framework named AAND. It comprises an Anomaly Amplification Stage to enhance the teacher model's discrimination capacity, and a Normality Distillation Stage to train the student model's normality reconstruction ability.
% In the first stage, we synthesize anomalies to aid training and introduce a novel RAA module to advance the teacher model. Specifically, the RAA module not only amplifies anomalies but also preserves the integrity of the pre-trained model. It includes a matching-guided residual gate and an attribute-scaling residual generator, which regulate the proportion and characteristics of the residuals, respectively. In the subsequent stage, a reverse distillation paradigm is employed to train a student decoder, in which a novel HKD loss is proposed to reconstruct complex normal patterns effectively. During inference, anomaly detection is facilitated by exploiting the discrepancies between the teacher and student models.

\subsection{{Revisiting Teacher-Student Discrepancy} } 
{
Knowledge Distillation (KD)~\cite{RD,RDplus,destseg} has achieved promising results in the industrial anomaly detection task. They solve the anomaly detection problem by focusing on the ``teacher-student discrepancy", \emph{i.e.}, the feature discrepancy between a fixed pre-trained teacher encoder and a learnable student decoder, which can be formulated as follows:
\begin{equation}\label{discre1}
D(\mathbf{F}_{T}, \mathbf{F}_{S}) = 1-s(\mathbf{F}_{T}, \mathbf{F}_{S}),
\end{equation} 
where $s(,)$ denotes the cosine similarity.  $\mathbf{F}_{T}$ and $\mathbf{F}_{S}$ represent the teacher features and student features, respectively.
It is hypothesized that anomalies
have larger discrepancies than normal samples as follows:
\begin{equation}\label{discre2}
D(\mathbf{F}_{T}^{a}, \mathbf{F}_{S}) > D(\mathbf{F}_{T}^{n}, \mathbf{F}_{S}),
\end{equation} 
where the superscript $n$ and $a$ represent normal and abnormal samples, respectively.
To make the hypothesis work, as shown in Fig.~\ref{fig1}, two underlying sub-assumptions are required in the recent KD-based frameworks.
    \textbf{(1) Assumption~I:} The teacher encoder can represent two separable distributions for normal and abnormal patterns, which is implicitly satisfied by employing an ImageNet pre-trained encoder.
   \textbf{(2) Assumption~II:} The student decoder is designed to reconstruct only the normal distribution, which is indirectly achieved by training exclusively on normal samples.
If the two assumptions were satisfied, the abnormal teacher features $\mathbf{F}_{T}^{a}$ would fall outside the boundary of the normal features  $\mathbf{F}_{T}^{n}$
, thereby exceeding the reconstruction capabilities of the student network $\mathbf{F}_{S}$. As a result, anomalies would exhibit larger discrepancies than normal samples, which would lead to a robust "teacher-student discrepancy" as illustrated in Eq.~\eqref{discre2}.}

{However, these
ideal assumptions are hard to meet in practice, which motivates us to enhance them from various perspectives.
As shown in Fig.~\ref{fig3}, we propose a simple yet effective two-stage anomaly detection framework, which comprises an Anomaly Amplification Stage (\textbf{Stage I}) to address Assumption I and a Normality Distillation Stage (\textbf{Stage II}) to address Assumption II. 
}
% In the first stage, we synthesize anomalies to aid training and introduce a novel RAA module to advance the teacher model. Specifically, the RAA module not only amplifies anomalies but also preserves the integrity of the pre-trained model. It includes a matching-guided residual gate and an attribute-scaling residual generator, which regulate the proportion and characteristics of the residuals, respectively. In the subsequent stage, a reverse distillation paradigm is employed to train a student decoder, in which a novel HKD loss is proposed to reconstruct complex normal patterns effectively. During inference, anomaly detection is facilitated by exploiting the discrepancies between the teacher and student models.

\begin{algorithm}[th] 
\SetAlgoLined %显示end
\caption{{Pseudo-code of AAND training}}\label{alg1}
\label{AL1}
% \KwIn{$T$: Teacher, $R$: RAA module, $A$: Advanced Teacher $B$: Bottleneck, $S$: Student decoder, subscript $k$: block index, superscript $a$: anomaly, $n$: normal}
% \tcc{Stage1: Anomaly Amplification}
$T$: Teacher, $R$: RAA module, $A$: Advanced Teacher $B$: Bottleneck, $S$: Student decoder, subscript $k$: block index, superscript $a$: anomaly, $n$: normal\\
\tcc{Stage1: Anomaly Amplification}
   \For{ $\mathbf{I}^a \leftarrow$ Synthesis Dataloader}{
$\mathbf{F}_{T_1}, \mathbf{F}_{T_2}, \mathbf{F}_{T_3}$ $\leftarrow T(\mathbf{I}^a)$; \\
  $w_k^a, \Delta_k^a \leftarrow R_k(\mathbf{F}_{T_k})$; // RAA module \\
 $\mathbf{F}_{A_k} \leftarrow \mathbf{F}_{T_k} + w_k^a\cdot \Delta_k^a$; 
\\
 $\mathcal{L}_1 
 \leftarrow \mathcal{L}_{focal}+\mathcal{L}_{A}$;  \\
 Optimize $R$ by $\mathcal{L}_1$;  \\}
\tcc{Stage2: Normality Distillation} 
\For{ $\mathbf{I}^n \leftarrow$ Train Dataloader}{
$\mathbf{F}^n_{A_1}, \mathbf{F}^n_{A_2}, \mathbf{F}^n_{A_3} \leftarrow $ Stage1;\\
$\mathbf{F}_{S_1}^n, \mathbf{F}_{S_2}^n, \mathbf{F}_{S_3}^n$ $\leftarrow 
S(B(\mathbf{F}_{A_1}^n, \mathbf{F}_{A_2}^n, \mathbf{F}_{A_3}^n))$;\\
$\mathcal{L}_2\leftarrow
\mathcal{L}_{KD}+\mathcal{L}_{HKD}$; \\
 Optimize $B$ and $S$ by $\mathcal{L}_2$; \\}
\end{algorithm}

% \subsection{Anomaly-Amplified Teacher Encoder}\label{3.1}
\subsection{Anomaly Amplification Stage}\label{3.1}
% An anomaly-amplified teacher encoder is proposed to extract multi-scale discriminative features for industrial images in this stage.
% In practice, some defective industrial products cannot be successfully detected, as the pre-trained embedding has limited discrimination capacity between anomalies and similar normal data. 
% Due to the gap between ImageNet pretraining and IAD task, it is challenging to guarantee the first assumption, \emph{i.e.}, jointly representing two different distributions for normal and abnormal patterns.
ImageNet pre-trained encoders are widely used \cite{patchcore,RD,RDplus} in the industrial anomaly detection task, where the pre-trained embedding is directly used as teacher representation without any adaption. However, as outlined in Table.~\ref{teacher_tab}, the gap between ImageNet pretraining and IAD task poses significant challenges in fulfilling~\textbf{Assumption I}. To mitigate this issue, we introduce an anomaly amplification stage to increase the teacher model's ability to discriminate between normal and anomalous patterns. Moreover, we propose an RAA module with adaptive residual generation to amplify anomaly and preserve the pre-trained model's generalizability. The training within this stage incorporates synthetic anomalies.

% \textcolor{red}{In unsupervised settings with only normal samples during training, anomaly amplification becomes non-trivial. Although synthetic anomalies can help train the teacher's discrimination ability, over-reliance can lead to catastrophic forgetting~\cite{kirkpatrick2017overcoming} of pre-trained knowledge. Therefore, we propose a novel Residual Anomaly Amplification (RAA) module, utilizing a residual learning mechanism to advance the vanilla teacher model. } 

% \textcolor{red}{For an image $\mathbf{I}$ $\in \mathbb{R}^{C\times H\times W}$, the original pre-trained teacher encoder is employed to extract $K$-level features $\mathbf{F}_{T_k}\in\mathbb{R}^{C_k\times H_k\times W_k}$ by $K$ sequential blocks. Our anomaly-amplified teacher encoder is then created by introducing $K$ residual-based anomaly amplification modules into the original encoder. 
%  The features extracted from our Advanced Teacher (A) network are denoted as $\mathbf{F}_{A_k}\in\mathbb{R}^{C_k\times H_k\times W_k}$. This training
% stage involves the synthetic anomalies.
% }

% To simplify the explanation, we omit the subscript k in the following discussion.
% the integrity of pre-trained embedding needs to be preserved, as it provides prior knowledge about capturing diverse normal patterns and being able to distinguish some unseen anomalies. 

 % The training of RAA module involves the exposure of synthetic anomalies.
\textbf{Anomaly synthesis.} 
 Given the normal images from the training set $\mathcal{X}^{train}$, we use a widely used anomaly generator~\cite{zavrtanik2021draem} to generate abnormal masks and images. First, Perlin noise \cite{perlin1985image} is generated and binarized into an anomaly mask $\mathbf{\overline{M}^a}$ $\in \{0,1\}^{H\times W}$ to determine the locations of anomalies. Then, texture samples are obtained from the source texture dataset \cite{cimpoi2014describing} to serve as the content for these locations.
For a normal image $\mathbf{I^n}$ $\in \mathbb{R}^{C\times H\times W}$, the corresponding synthetic abnormal image is denoted as $\mathbf{I^a}$ $\in \mathbb{R}^{C\times H\times W}$. Considering the anomalies are typically on the foreground object, we follow~\cite{FOD} to extract a foreground
mask $\mathbf{\overline{M}^f}$ $\in \{0,1\}^{H\times W}$ with grayscale binary thresholding algorithms. The resulting pseudo-anomaly mask $\mathbf{\overline{M}}$ is then obtained by taking the intersection of the two masks.

During the first-stage training, we take the corrupted image $\mathbf{I^a}$ as input of the teacher encoder. The encoder extracts $K$-level features 
$\mathbf{F}_{T_k}\in\mathbb{R}^{C_k\times H_k\times W_k}$. 
At the k-th level of feature extraction, a total of $H_k \times W_k$ patches are obtained. By subsampling the anomaly mask at the same scale, denoted as $\mathbf{\overline{M}}_k$, and using it to label the normal and abnormal locations, these patches can be divided into $N_k^n$ normal patches and $N_k^a$ abnormal patches. The corresponding features are denoted as $\mathbf{F}_{T_k}^n\in\mathbb{R}^{N^n_k\times C_k}$ and $\mathbf{F}_{T_k}^a\in\mathbb{R}^{N^a_k\times C_k}$ respectively. 

% \textcolor{red}{With the exposure of synthetic anomalies,
% we then design a novel residual anomaly amplification module, which can enhance
% the discrimination capacity of the vanilla teacher encoder.}
% Due to the gap between synthetic anomalies and real-world anomalies, directly using them to train the teacher model may lead to catastrophic forgetting~\cite{kirkpatrick2017overcoming} of pre-trained knowledge. Therefore, we propose a novel Residual Anomaly Amplification (RAA) module, which utilizes a residual learning mechanism to advance the vanilla teacher model. }

\begin{figure}[t]
    \centering
\includegraphics[scale=0.335]{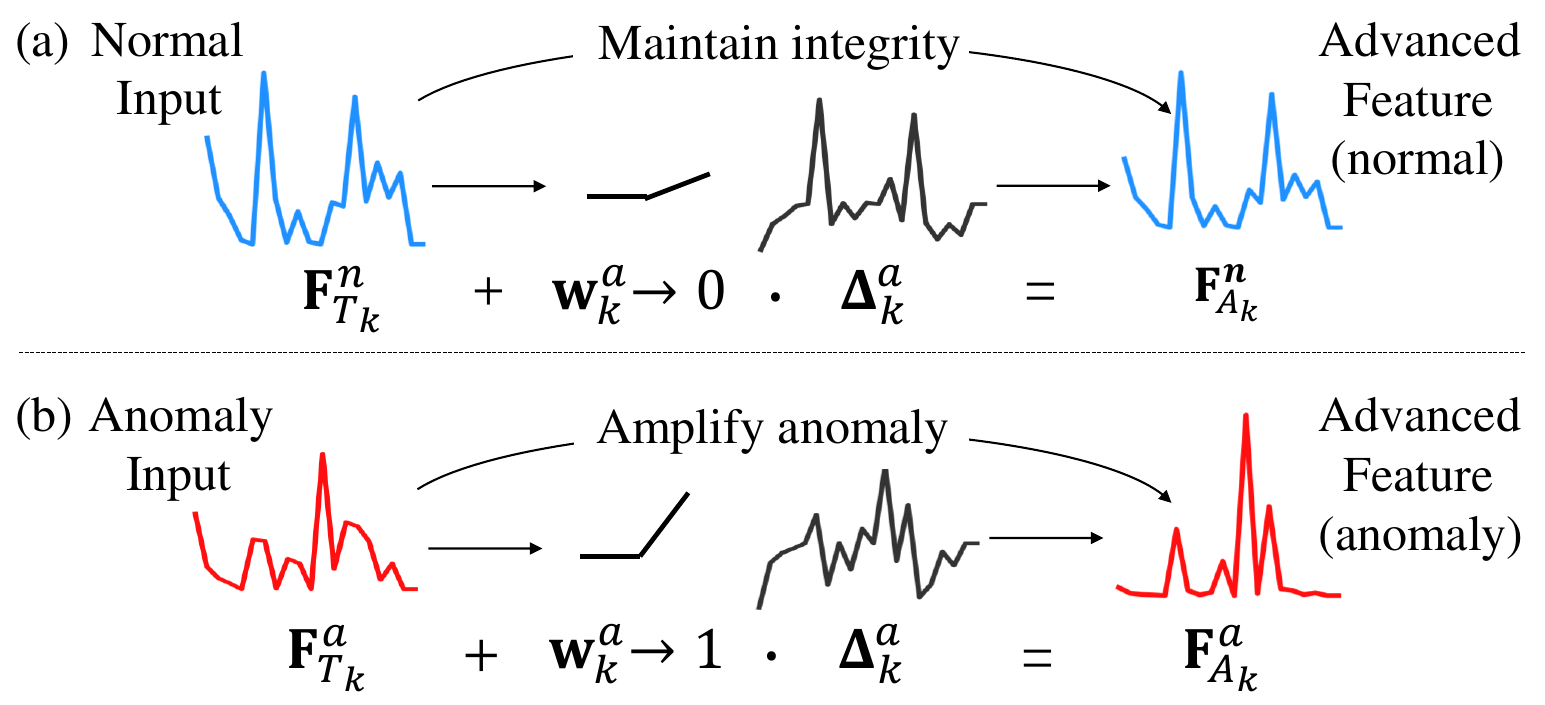}
    \caption{Illustration of our proposed RAA module. It utilizes an adaptive residual learning mechanism to amplify anomalies while maintaining the integrity of pre-trained model.}
    \label{RAA}
\end{figure}

\textbf{Residual Anomaly Amplification.} 
Considering over-reliance on synthetic
anomalies may lead to catastrophic forgetting~\cite{forget} of pre-trained knowledge, we propose a novel RAA module, which utilizes an adaptive residual learning mechanism to enhance anomaly detection without compromising the integrity of the pre-trained model.
This module comprises two critical components: a matching-guided residual gate and an attribute-scaling residual generator. As shown in Fig.~\ref{RAA}, the residual gate suppresses residuals on normal samples to maintain
the integrity of pre-trained model while encouraging residuals on synthetic
inputs to promote deviations. Additionally, the residual generator is designed to adaptively learn the characteristics of residuals, which directly influence the distribution of emergent abnormal features. By embedding these RAA modules into the pre-existing framework, we have successfully crafted an advanced teacher encoder that robustly enhances anomaly detection capabilities.
 The extracted feature at the k-th block of the advanced teacher model, denoted as $\mathbf{F}_{A_k}\in\mathbb{R}^{C_k\times H_k\times W_k}$, is calculated as follows: 
% \begin{equation}
% \mathbf{F}_{A_k}= \mathbf{F}_{T_k}+\mathrm{G}_{\theta_1}(\mathbf{F}_{T_k})\cdot \mathrm{R}_{\theta_2}(\mathbf{F}_{T_k}),
% \end{equation}
% where $\mathrm{G}_{\theta_1}(\mathbf{F}_{T_k})$ is the residual gate, which produces anomaly weights $\mathrm{w}^a \in [0,1]^{H_k\times W_k}$. Besides, $\mathrm{R}_{\theta_2}(\mathbf{F}_{T_k})$ is the residual generator, which generates residuals $\mathbf{\Delta}^a \in \mathbb{R}^{C_k\times H_k\times W_k}$. The parameters $\theta_1$ and $\theta_2$ correspond to their respective network. 
\begin{equation}
\mathbf{F}_{A_k}= \mathbf{F}_{T_k}+\mathbf{w}^a_k\cdot \mathbf{\Delta}_k^a,
\end{equation}
where $\mathbf{w}_k^a \in [0,1]^{H_k\times W_k}$ represents the anomaly weights predicted by the residual gate, and $\mathbf{\Delta}_k^a \in \mathbb{R}^{C_k\times H_k\times W_k}$ is the residual signals generated by the residual generator.

% For simplicity, we omit the subscript $k$ in the following discussion.  
% The residual generation manner amplifies the anomalies while leveraging the original discrimination capacity of the pre-trained model, which will be further explained.
% The variable $k$ represents the k-th level of feature extraction. For simplicity, we omit the subscript $k$ in the following discussion.
% \newpage

% This module functions similarly to a binary classifier but has two significant differences. Firstly, the anomaly weight is assigned in a matching-based manner, where the normal memories and
% anomaly memories encourage the diversity of normality and anomalies. Secondly, the anomaly weight is not utilized directly for anomaly detection. Instead, its purpose lies in regulating the proportion of residuals to suppress residuals on normal data while encouraging residuals on anomalous inputs.

\subsubsection{Matching-guided residual gate}
This module aims to regulate the proportion of residuals, which functions as a robust binary classifier that predicts the weight of residuals. This prediction is informed by the matching results between the queries and the learned memory items.
  
As illustrated in Fig.~\ref{fig3}, we take the feature map $\mathbf{F}_{T_k}\in\mathbb{R}^{C_k\times H_k\times W_k}$ extracted from the frozen teacher block as input. The input is first added with a position encoding $\mathbf{p}_c$ \cite{csflow}, and then passed through an MLP, $\Phi_{\theta_1}^k(\cdot)$,  for projection:
\begin{equation}\label{reform}
\mathbf{q}_k^i = \Phi_{\theta_1}^k(\mathbf{F}_{T_k}^i+\mathbf{p}_c^i),
\end{equation}
where $\mathbf{F}_{T_k}^i\in \mathbb{R}^{C}$ and $\mathbf{p}_c^i\in \mathbb{R}^{C}$ are the input feature and position encoding at the $i$-th patch, respectively. In addition, $\mathbf{q}_k^i\in \mathbb{R}^{C}$ represents the projected feature.
To represent the memory of normality and anomaly, we configure a normal memory $\mathbf{M}_k^n=\{\mathbf{m}_{k1}^n, \mathbf{m}_{k2}^n, ...,\mathbf{m}_{kL}^n\}$ containing $L$ learnable normal embeddings, and an anomaly memory $\mathbf{M}^a_k= \{\mathbf{m}_{k1}^a, \mathbf{m}_{k2}^a, ...,\mathbf{m}_{kL}^a\}$ containing $L$ learnable anomaly embeddings. {These embeddings are initialized with a Gaussian distribution $\mathcal{N}(\mathbf{0},\mathbf{1})$.}
Subsequently, a matching process is conducted between the query and memories, which involves computing their cosine similarity and normalizing the similarity scores using the softmax function. Consequently, the normal weights, denoted as $w^n_{kj}, j \in \{1, 2, ..., L\}$, and the anomaly weights, denoted as $w^a_{kj}, j \in \{1, 2, ..., L\}$, are derived as follows:
\begin{equation}\label{w}
\begin{aligned}
w^n_{kij}=\frac{\mathrm{exp}(s(\mathbf{q}_k^i,\mathbf{m}^n_{kj}))}
{Z}, &\,  w^a_{kij}=\frac{\mathrm{exp}(s(\mathbf{q}_k^i,\mathbf{m}^a_{kj}))}
{Z},
\end{aligned}
\end{equation}
where $Z$ represents the normalization factor, ensuring that $\sum_{j=1}^L w^n_{kij}+ \sum_{j=1}^L w^a_{kij}=1$. 
When presented with an abnormal patch query $\mathbf{q}_k^a$, it has a higher affinity to the anomaly memory. Conversely, for a normal patch query $\mathbf{q}_k^n$, it
has a higher affinity to the normal memory.  
Afterwards, the sum of anomaly weights is taken as the outputs of the residual gate:
\begin{equation}\label{w}
w_{ki}^a = \sum_{j=1}^L w^a_{kij}=1-\sum_{j=1}^L w^n_{kij},
\end{equation}
where the anomaly weight $w^a_{ki} \in [0,1]$ represents the anomaly probability of inputs. To supervise the anomaly weight, we adopt the well-known focal loss~\cite{lin2017focal}. The ground truth $\overline{w}^a_{ki}\in\{0,1\}$ is from the pseudo-anomaly mask $\mathbf{\overline{M}}_k$, which equals to 1 when it is a synthetic abnormal patch, while equaling to 0 when it is a normal patch. {Under
binary label supervision, the discriminative ability of the residual gate is guaranteed. As for diversity, by matching with varied normal and synthetic abnormal features, the dual memories will implicitly learn the diverse feature distributions. Within the RAA module, the role of memory is to control the proportion of adaptive residual
generation for a robust teacher-student discrepancy. Given that the module constitutes only a part of our
AAND framework, we do not impose further constraints to maintain its simplicity.}
\subsubsection{Attribute-scaling residual generator}
Due to the influence of residual characteristics on the distribution of new abnormal features, we address this challenge by introducing an attribute-scaling residual generator.
The residual noise aims to emphasize the
attributes that are relevant to anomalies while diminishing those irrelevant attributes. Specifically, the
key designs are the channel weight module and the anomaly
amplification loss.

As illustrated in Fig.~\ref{fig3}, we initiate the process by projecting $\mathbf{F}_{T_k}^i$ using another MLP, denoted as $\Phi_{\theta_2}^k(\cdot)$.  Subsequently, each channel of the projected feature is mapped to the range (-1,1) through a tanh function, yielding scaling weights $\mathbf{\Sigma}^i_k\in\mathbb{R}^C$ for each channel. Finally, the generated residual $\mathbf{\Delta}_k^{a_i}\in\mathbb{R}^C$ is calculated by multiplying the channel weights and input features as follows:
\begin{equation}\label{delta}
\mathbf{\Delta}_k^{a_i} = \mathrm{Tanh}(\Phi_{\theta_2}^k(\mathbf{F}_{T_k}^i))\odot  \mathbf{F}_{T_k}^i,
\end{equation}
where $\odot$ denotes element-wise multiplication.
To obtain the updated features and avoid interfering with the parameters of the residual gate, we replace the predicted anomaly weights $\hat{w}_i^a$ with the ground truth anomaly weights $\overline{w}_i^a$ during training as follows:
% \begin{equation}
% \begin{aligned}
% & \mathbf{F}_{A}^{n_i} = \mathbf{F}_{T_k}^{n_i} + 0\cdot \mathbf{\Delta}_i \\
% & \mathbf{F}_{A}^{a_i} = \mathbf{F}_{T_k}^{a_i} + 1\cdot \mathbf{\Delta}_i. 
% \end{aligned}
% \end{equation}
\begin{equation}\label{A_train}
 \mathbf{F}_{A_k}^{n_i} = \mathbf{F}_{T_k}^{n_i} + 0\cdot \mathbf{\Delta}^{a_i}_{k}, \quad \mathbf{F}_{A_k}^{a_i} = \mathbf{F}_{T_k}^{a_i} + 1\cdot \mathbf{\Delta}^{a_i}_{k},
\end{equation}
% Here, $\mathbf{F}_{A}^{n_i}$ and $\mathbf{F}_{A}^{a_i}$ represents the features for the i-th normal patch and abnormal patches extracted by the advanced teacher, whereas  denotes the features for abnormal patches during training. 
where the superscripts $a_i$ and $n_i$
denote the abnormal and normal patches, respectively. Here, $\mathbf{F}_{A_k}^{a_i}$ represents the learnable features updated by the residuals, while $\mathbf{F}_{T_k}^{a_i}$, $\mathbf{F}_{A_k}^{n_i}$, and $\mathbf{F}_{T_k}^{n_i}$ remain frozen. This ensures that the optimization is confined exclusively to the synthetic anomaly features, specifically on the residual $\mathbf{\Delta}^{a_i}_{k}$.

To supervise the residuals, we propose an \textbf{anomaly amplification loss} to push the abnormal features outside the boundaries of normality. {Unlike
traditional contrastive learning employing both ”push” and ”pull” actions, our method focuses solely
on pushing anomalies away from normal samples using a dynamic margin, enhancing discriminability
while preserving pre-trained priors.} The loss is calculated as follows:
\begin{equation}\label{laa}
\begin{aligned}
\mathcal{L}_{\mathrm{A}} = 
\frac{1}{N^a_kN^n_k}\sum\limits_{i=1}^{N^a_k}\sum\limits_{j=1}^{N^n_k} \mathrm{max}(s(\mathbf{F}^{a_i}_{A_k},\mathbf{F}^{n_j}_{T_k}),\mathbf{S}^{ref}_{kij}),\\
\end{aligned}
\end{equation}
where
$\mathbf{S}^{ref}_k\in \mathbb{R}^{N^a_k\times N^n_k}$ is a dynamic margin matrix calculated based on the similarity matrix produced by the vanilla teacher model, which is denoted as follows: 
\begin{equation}
\mathbf{S}_{kij}^{ref}=s(\mathbf{F}^{a_i}_{T_k}, \mathbf{F}^{n_j}_{T_k})-\alpha,
\end{equation}
where $\alpha$ is a hyper-parameter to control the degree of reduction. It is noteworthy that the loss function $\mathcal{L}_{A}$ is specifically designed to reduce the similarity between abnormal and normal features. For normal samples, we utilize the residual gate to suppress the residuals instead of pulling them as in standard contrastive learning. This approach preserves the integrity of the pre-trained model, thereby preserving its ability to capture a diverse representation of normal samples.

%  \textbf{Training. }
%  The focal loss and anomaly amplification loss supervise the anomaly weight $\mathbf{w}^a_k$ and the residual signals $\mathbf{\Delta}^a_k$, respectively. They optimize the parameters of the MRG and ARG independently, without causing interference between them. 

The overall loss of this stage is summarized as follows:
 \begin{equation}
\mathcal{L}_1 
= \mathcal{L}_{focal}+\mathcal{L}_{A},
 \end{equation}
where the focal loss and anomaly amplification loss supervise the anomaly weight $\mathbf{w}^a_k$ and the residual signals $\mathbf{\Delta}^a_k$, respectively. ~{Under the supervision,
the abnormal features are trained to be pushed outside the boundaries of normality.
As a result, Stage I enhances Assumption I, \emph{i.e.}, the teacher model can represent two separable distributions for the normal and abnormal
patterns.}

{It should be noted that even if Stage I provides a good foundation, the ``teacher-student discrepancy" of unseen anomalies is still hard to guarantee. Thus, we need Stage II to model the distribution of normal features, allowing anomalies to be naturally identified as deviations from the learned distribution.}
% They optimize the parameters of \textcolor{red}{the MRG and ARG independently, without causing interference between them. }

 % For positive pairs, \emph{i.e.}, the similarity between normal samples, we do not perform optimization, but instead use the residual gate to preserve the integrity of the pre-trained model, which characterizes the diverse representation of normal samples.

% \subsection{Normality-recalled Student Decoder}\label{3.4}
\subsection{Normality Distillation Stage}\label{3.2}
Upon enhancing the discrimination capabilities of the teacher encoder, our objective shifts to training a student decoder, designed to reconstruct only the normal distribution. 
To accomplish this, we employ a reverse distillation paradigm~\cite{RD} to train the student decoder. Within this framework, we introduce a hard knowledge distillation loss, specifically designed to improve the reconstruction of normal patterns.

% The student distills the representation of the advanced teacher on only normal samples, 
% while failing to reconstruct
% the abnormal embedding from teacher model. 

In this stage, the advanced teacher is fixed and only receives normal samples. Its output normal features $\mathbf{F}^{n}_{A_k}\in\mathbb{R}^{C_k\times H_k\times W_k}$ are then passed to a one-class bottleneck embedding module~\cite{RD}, which aggregates the multi-scale embeddings to obtain a compact representation. Subsequently, the learnable student decoder predicts multi-scale features $\mathbf{F}^{n}_{S_k}\in\mathbb{R}^{C_k\times H_k\times W_k}$, which is trained to fit the target features from the teacher model using a knowledge distillation loss:
\begin{equation}\label{kd}
  \mathcal{L}_{KD} =\ 1-s(\mathbf{\Tilde{F}}^{n}_{A_k},\mathbf{\Tilde{F}}^{n}_{S_k}),
\end{equation}
where the features $\mathbf{\Tilde{F}}^{n}_{A_k}\in\mathbb{R}^{C_kH_kW_k}$ and $\mathbf{\Tilde{F}}^{n}_{S_k}\in\mathbb{R}^{C_kH_kW_k}$ are obtained by flattening the features $\mathbf{F}^{n}_{A_k}$ and $\mathbf{F}^{n}_{S_k}$ through vectorization, respectively.

\begin{figure}[t]
    \centering
\includegraphics[scale=0.60]{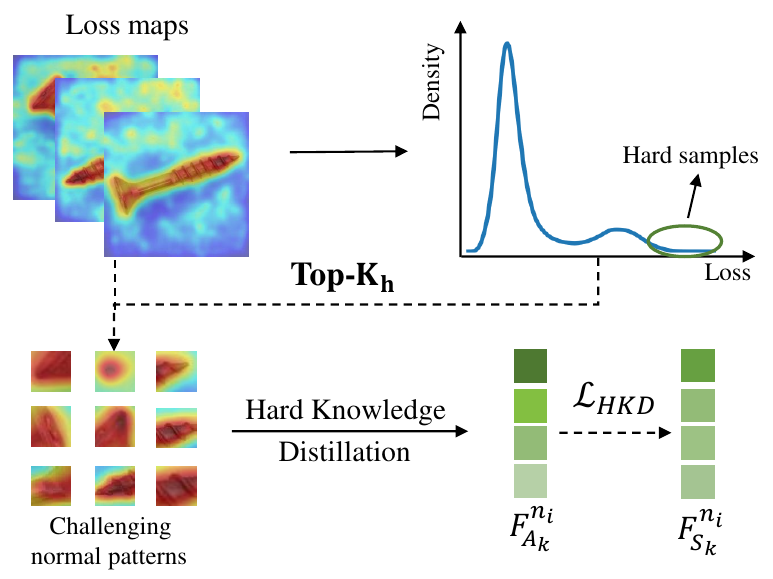}
    \caption{Illustration of our proposed HKD loss. It facilitates the reconstruction on challenging normal patterns by selecting the $K_h$ normal patches with the highest distillation loss for further training.}
    \label{HKD}
\end{figure}

Furthermore, considering the difficulty in accurately reconstructing the features of certain fine-grained normal textures and rare normal patterns,
we propose a hard knowledge distillation loss as follows:
% \begin{equation}\label{kd}
% \begin{aligned}
%   &\mathcal{L}_{KD} =\ 1-s(\mathbf{\Tilde{F}}^{n}_{A_k},\mathbf{\Tilde{F}}^{n}_{S_k})\,  \\
%   &\mathcal{L}_{H} =\, \frac{1}{K_h}\sum\nolimits_{i=1}^{K_h}(1-s(\mathbf{F}^{n_i}_{A_k}, \mathbf{F}^{n_i}_{S_k}))\\
% &\mathcal{L}_{HKD}=\mathcal{L}_{KD}+\mathcal{L}_{H},
% \end{aligned}
% \end{equation}
\begin{equation}\label{hkd}
\mathcal{L}_{HKD} =\, \frac{1}{K_h}\sum_{d=1}^{K_h}(1-s(\mathbf{F}^{n_d}_{A_k}. \mathbf{F}^{n_d}_{S_k})),
\end{equation}
where $K_h$ represents the number of selected hard samples, while $\mathbf{F}^{n_d}_{A_k}$ and $\mathbf{F}^{n_d}_{A_k}$ denote the selected features. As illustrated in Fig.~\ref{HKD}, the HKD loss specifically emphasizes the distillation of the top-$K_h$ hard normal instances, which helps the student decoder to achieve a more accurate reconstruction of these challenging patterns.

The overall loss function for the normality distillation stage is formulated as follows:
\begin{equation}\label{l2}
\mathcal{L}_{2} = \mathcal{L}_{KD}+\mathcal{L}_{HKD}.
\end{equation}

% While some recent methods~\cite{destseg,RDplus}
% enforce stronger constraints on the student network, they make the framework more complex by introducing additional network blocks and multiple loss functions. For the relative simplicity and satisfactory performance of RD~\cite{RD}, we adopt it as our baseline instead of employing these more complex methods.
% This decision is based on its relative simplicity and satisfactory performance, making it preferable over more complex alternatives.

% The robust teacher-student feature discrepancy between the two networks effectively facilitates the detection of anomalies.
{By exclusively training with normal samples and leveraging the compact embedding design of reverse distillation~\cite{RD}, the student decoder will
be effectively constrained to reconstruct only the normal
distribution. With the HKD loss, the normality recall ability towards challenging normal patterns is further strengthened.  As a
result, Stage II enhances Assumption II, \emph{i.e.}, the student decoder can only reconstruct the normal
distribution.}
To summarize the two-stage training procedure of our  
AAND, we show the pseudo-code in Algorithm \ref{alg1}.

\subsection{Inference}\label{inference}
During inference, the anomaly detection is based on the feature discrepancy between the advanced teacher model and the student model. Given a test image $\mathbf{I}^t$, the advanced teacher encoder extracts $K$-level features  $\mathbf{F}_{A_k}$, then the student decoder aggregates the features and reconstructs the multi-scale features $\mathbf{F}_{S_k}$. The feature discrepancy, denoted as $D(\mathbf{F}_{A_k}, \mathbf{F}_{S_k})$, is calculated using the negative cosine similarity as follows:
\begin{equation}\label{eq1}
D(\mathbf{F}_{A_k}, \mathbf{F}_{S_k}) = 1-s(\mathbf{F}_{A_k}, \mathbf{F}_{S_k}).
\end{equation} 

The advanced teacher model is capable of generating discriminative features for both normal and abnormal inputs, whereas the student network is tailored to reconstruct only the normal features. This configuration results in a minor feature discrepancy in normal samples, while a pronounced discrepancy when anomalies are present. To effectively localize anomalies, anomaly maps are generated across the $K$ network blocks. These maps are then upsampled and aggregated to obtain an anomaly score for each pixel:
\begin{equation}\label{eq8}
\mathbf{\hat{S}}_A = \sum_{l=1}^{K} \mathrm{Up}_k(D(\mathbf{F}_{A_k}, \mathbf{F}_{S_k})),
\end{equation}
 where $\mathbf{\hat{S}}_A$ is the predicted anomaly score map, and $\mathrm{Up}_k(\cdot)$ denotes the bi-linear up-sampling.   A higher score in the anomaly map indicates a higher likelihood of anomaly at that position, and the maximum value in the anomaly map is considered as the image-level anomaly score.

\section{Experiments}\label{exper}
\subsection{Dataset}
To assess the effectiveness of the proposed method, we conduct comprehensive experiments on three datasets: MVTec AD \cite{mvtec}, Visa \cite{visa} and MVTec3D-RGB \cite{mvtec3d}.

\textbf{MVTec AD.} It is a well-studied benchmark for anomaly detection and localization. It consists of 3,629 anomaly-free training images, and 1,725 testing images, which include both normal and abnormal samples. The dataset has been widely explored with the results almost reaching saturation.

\textbf{VisA.} It is a large industrial anomaly detection dataset
that is composed of 10,821 high-resolution color images of 12 objects. The anomaly type covers both surface
and structural defects. The dataset is more challenging than MvTec AD. It not only contains multiple instances of a single object category but also has different types of Printed Circuit Boards~(PCB) with complex structures.

\textbf{MVTec3D-RGB.} It is a multi-modal dataset that comprises 2,656 training 2D-3D pairs and 1,197 testing 2D-3D pairs. Some anomalies are more significant in geometry while having subtle changes in appearance, making it particularly challenging to detect anomalies based solely on the visual appearance of RGB images. Considering the computational cost with point cloud data, we follow the recent work~\cite{FOD} to employ only the RGB images for anomaly detection and localization.

\begin{table*}[t]
\setlength\tabcolsep{6.8pt}
% \scriptsize
\centering
\caption{Anomaly detection and localization performance on the MVTec AD~\cite{mvtec}, VisA~\cite{visa} and MvTec3D-RGB~\cite{mvtec3d} dataset. \textbf{Since the MVTec AD benchmark has approached saturation, our method shows only slight enhancements.} When applied to the more challenging VisA and MVTec3D-RGB datasets, more significant improvements can be observed. Compared to P-AUC, the PRO metric offers a more effective reflection of the model's ability to detect hard samples.}
\label{tab1}
\begin{tabular}{c|c|c|ccc|ccc|ccc}
  % \hline
  \toprule
& \multirow{2}*{Method}& \multirow{2}*{Venue}&  \multicolumn{3}{c|}{MvTec AD}&\multicolumn{3}{c|}{VisA}&\multicolumn{3}{c}{MvTec3D-RGB}\\
 % \cmidrule(r){3-5} \cmidrule(r){6-8}
 &&& I-AUC&P-AUC&PRO& I-AUC&P-AUC&PRO& I-AUC&P-AUC&PRO\\
    % \hline
    \midrule
\multirow{5}*{\rotatebox{90}{Others}} 
&DRAEM~\cite{zavrtanik2021draem}&ICCV'21&98.0&97.3&-&88.7&93.5&72.4&75.7&97.6&-\\
&PaDiM~\cite{padim}&ICPR'21&95.5&97.3&92.1&89.1&98.1&85.9&76.4&-&-\\
&Patchcore~\cite{patchcore}&CVPR'22&99.1&98.1&94.4&95.1&98.8&91.2&77.0&96.6&87.6\\
&CFLow~\cite{cflow}&WACV'22&98.3&98.0&94.6&93.2&98.4&89.0&85.1&97.4&-\\
&SimpleNet~\cite{liu2023simplenet}&CVPR'23&\bf{99.6}&98.1&-&80.8&89.5&69.4&84.1&96.0&88.0\\
&FOD~\cite{FOD}&ICCV'23&99.2&98.3&-&93.7&98.4&-&88.4&97.6&-\\
% \hline
\midrule
\multirow{4}*{\rotatebox{90}{KD}}
&ST~\cite{ST} &BMVC'21&95.5&97.0&92.1&83.3&-&62.0&-&-&-\\
&RD~\cite{RD} (baseline)&CVPR'22&98.5&98.3&94.0&96.2&98.6&93.4&88.5&99.0&96.7\\
&RD++~\cite{RDplus}&CVPR'23&99.4&98.3&95.0&96.0&98.6&93.2&88.2&99.0&97.2\\
% &MKD\cite{MK}&&&&&&\\ 
&DeSTSeg~\cite{destseg}&CVPR'23&98.6&97.9&-&91.2&98.0&90.2&86.3&96.7&89.8\\
&\bf{Ours}&&99.5&\bf{98.4}&\bf{95.0}&\cellcolor{gray!20}\bf{98.1}&\cellcolor{gray!20}\bf{99.0}&\cellcolor{gray!20}\bf{94.8}&\cellcolor{gray!20}\bf{90.6}&\cellcolor{gray!20}\bf{99.2}&\cellcolor{gray!20}\bf{97.5}\\
   \bottomrule
   % \bottomrule
\end{tabular}
\end{table*}

\begin{table*}[ht]
\setlength\tabcolsep{5.5pt}
\centering
\caption{{Evaluation of anomaly detection and localization performance of all object categories on the challenging VisA dataset.}}
\label{tab2}
% \vspace{-6pt} 
% \subtable[Performance of all object categories on the VisA dataset.]{
\begin{tabular}{c|c|cccccccccccc|c}
   \toprule
     &\multirow{2}*{Method}&\multirow{2}*{candle}&\multirow{2}*{capsules}&\multirow{2}*{cashew}&chewing&\multirow{2}*{fryum}&maca-&maca-&\multirow{2}*{pcb1}&\multirow{2}*{pcb2}&\multirow{2}*{pcb3}&\multirow{2}*{pcb4}&pipe&\multirow{2}*{\bf{mean}}\\    &&&&&gum&&roni1&roni2&&&&&fryum&\\
     \midrule
\multirow{3}*{\rotatebox{90}{I-AUC}} 
% Patchcore&\bf{98.6}&81.6&97.3&99.1&96.2&97.5&78.1&\bf{98.5}&97.3&97.9&99.6&99.8&95.1\\
&RD&95.3&90.7&97.0&99.1&96.5&96.5&90.6&97.0&96.9&95.5&99.5&99.5&96.2\\
&RD++&96.0&91.4&97.0&99.1&95.5&97.0&88.9&97.1&96.2&94.9&99.5&99.6&96.0\\
% \cline{3-14}
&\bf{Ours}&\bf{98.4}&\bf{92.3}&\bf{98.0}&\bf{99.7}&\bf{98.0}&\bf{99.6}&\bf{95.2}&\bf{98.4}&\bf{98.9}&\bf{98.7}&\bf{99.9}&\bf{99.9}&\bf{98.1}\\
% \midrule
% \multirow{3}*{\rotatebox{90}{P-AUC}} 
% &RD&98.8&\bf{99.5}& 95.0& 98.4& 96.9 &99.8 &99.6 &99.8& 98.9& 99.3 &98.4& 99.1&98.6\\
% &RD++&98.8&99.4&95.4&98.5&96.1&99.7&99.4&99.8&98.8&99.2&98.7&99.1&98.6\\
% % \cline{3-14}
% &\bf{Ours}&\bf{99.4}&99.3&\bf{97.0}&\bf{99.0}&\bf{97.0}&\bf{99.8}&\bf{99.7}&\bf{99.9}&\bf{99.2}&\bf{99.3}&\bf{98.8}&\bf{99.1}&\bf{99.0}\\
\midrule
\multirow{3}*{\rotatebox{90}{P-PRO}} 
&RD&93.5& 95.6& 91.3& 85.2& 92.4 &96.1& 98.0 &95.6& 92.4&94.8& 88.6 &96.7 &93.4\\
&RD++&92.4&94.9&90.9&84.9&89.7&94.9&97.2&95.2&91.4&94.3&91.0&96.2&92.8\\
% \cline{3-14}
&\bf{Ours}&\bf{95.5}&\bf{96.9}&\bf{94.8}&\bf{88.1}&\bf{92.9}&\bf{96.4}&\bf{98.5}&\bf{96.6}&\bf{94.4}&\bf{95.6}&\bf{91.4}&\bf{97.0}&\bf{94.8}\\
    \bottomrule
\end{tabular}
\end{table*}

\begin{table*}[ht]
\setlength\tabcolsep{6.6pt}
\centering
\caption{{Evaluation of detection and localization performance of all object categories on the challenging MVTec3D-RGB datasets.}}
\label{tab3}
% \subtable[Performance of all object categories on the MVTec3D-RGB dataset.]{
\begin{tabular}{c|c|cccccccccc|c}
   \toprule
     &\multirow{2}*{Method}&\multirow{2}*{bagel}&cable&\multirow{2}*{carrot}&\multirow{2}*{cookie}&\multirow{2}*{dowel}&\multirow{2}*{foam}&\multirow{2}*{peach}&\multirow{2}*{potato}&\multirow{2}*{rope}&\multirow{2}*{tire}&\multirow{2}*{\bf{mean}}\\
    &&&gland&&&&&&&&&\\
     \midrule
     \multirow{3}*{\rotatebox{90}{I-AUC}}&RD&98.5&96.2&96.5&64.5&99.1&83.6&91.4&68.0&99.4&87.7&88.5\\
&RD++&\bf{99.4}&91.5&94.4&69.7&98.6&82.6&92.7&69.5&97.7&85.5&88.2\\
&\bf{Ours}&99.3&\bf{96.3}&\bf{96.9}&\bf{71.2}&\bf{99.3}&\bf{86.8}&\bf{93.0}&\bf{71.2}&\bf{99.5}&\bf{92.7}&\bf{90.6}\\
% \midrule
% \multirow{3}*{\rotatebox{90}{P-AUC}} 
% &RD&99.2& 99.6& 99.6 &97.9 &99.6& 96.3& 99.5& 99.3& 99.6& 99.5&99.0\\
% &RD++&99.4&99.2&99.5&98.4&99.6&96.0&99.4&99.5&99.5&99.4&99.0\\
% % \cline{3-14}
% &\bf{Ours}&\bf{99.5}&\bf{99.7}&\bf{99.6}&\bf{98.7}&\bf{99.7}&\bf{96.5}&\bf{99.5}&\bf{99.4}&\bf{99.6}&\bf{99.7}&\bf{99.2}\\
 \midrule
\multirow{3}*{\rotatebox{90}{PRO}} 
&RD&97.4 &99.4 &99.0& 91.7& 99.0 &89.1 &\bf{98.6}& 98.2& 97.9& 98.8&96.9\\
&RD++&98.4&98.8&98.9&\bf{94.4}&99.1&90.3&98.4&98.3&97.0&98.3&97.2\\
% \cline{3-14}
&\bf{Ours}&\bf{98.4}&\bf{99.5}&\bf{99.2}&93.7&\bf{99.3}&\bf{90.3}&98.4&\bf{98.3}&\bf{98.7}&\bf{99.3}&\bf{97.5}\\
    \bottomrule
\end{tabular}

\end{table*}

\subsection{Implementation Details}
\textbf{Experimental settings.} We utilize the WideResNet50~\cite{wideresnet} encoder as the backbone with the output embeddings from three layers ($K=3$). During training, images are resized into 256 × 256, batch size is set to 16, and Adam is used as the optimizer with a learning rate of 0.005. We train the first stage for 100 epochs and then train the second stage for another 120 epochs. 
Besides, the hyper-parameters $\alpha$, $L$, and $K_h$ are set to 0.3, 50, and 10, respectively. For the synthetic anomalies, we follow \cite {BGAD} to extract foreground masks with grayscale binary thresholding algorithms to control the noise only in the foreground, which brings the synthesis closer to real conditions.

\textbf{Evaluation metrics.} We evaluate the AD performance using the area under the receiver operator curve (AUROC) based on the generated anomaly scores. We utilize image-level AUROC (I-AUC) to evaluate anomaly detection performance, while pixel-based AUROC (P-AUC) and the PRO metric~\cite{pro} were employed to assess anomaly localization. {Unlike P-AUC, the PRO metric is not influenced by the size of the anomaly regions. As a result, it can more effectively reflect the model's performance in detecting hard samples with subtle anomalies. }

\subsection{Main Results}
We conduct a comprehensive comparison of the performance of anomaly detection and localization. 
We mainly compare the Knowledge Distillation-based (KD) methods~\cite{ST,RD,destseg,RDplus} and other popular paradigms for anomaly detection, such as reconstruction-based approaches~\cite{mvtec,FOD}, synthesis-based methods~\cite{zavrtanik2021draem,liu2023simplenet}, and embedding-based methods~\cite{patchcore,cflow}.
For the MvTec AD dataset \cite{mvtec}, reported results are directly obtained from the published papers. For the newly published VisA \cite{visa} and MVTec3D-RGB \cite{mvtec3d} datasets, due to some anomaly detection methods have not yet been evaluated on these datasets, we re-train them using their open-source code and evaluate their performance on both datasets.

Table \ref{tab1} shows the
comparison results of anomaly detection and localization
on the MVTec AD, VisA, and MVTec3D-RGB benchmarks. In the widely studied MVTec AD dataset, our method slightly outperforms the current state-of-the-art methods.
Since the anomalies of this dataset are relatively easier to identify, the current methods have already achieved a high level of performance, leaving limited potential for substantial advancements.
In the VisA and MVTec3D datasets, our method surpasses the current state-of-the-art approaches significantly. Compared to the well-known method PatchCore~\cite{patchcore}, our method takes advantage of the enhanced feature encoder and robust knowledge distillation paradigm, leading to better performances.
Compared to the synthesis-based method SimpleNet~\cite{liu2023simplenet}, our approach achieves comparable results on the MvTec dataset, while demonstrating significant improvements on the more challenging Visa and Mvtec3D-RGB datasets. This observation indicates the robustness of our method in handling more complex industrial datasets.
Compared with our baseline RD, our proposed method improves the image-level AUROC and pixel-level PRO by 1.0\%, 1.0\% on MvTec AD, 1.9\%, 1.4\% on VisA, 2.1\%, 0.8\% on Mvtec3D-RGB. 

Thanks to the enhanced discrimination capability of our advanced teacher model, our method demonstrates the ability to effectively detect previously undetectable anomalies.
Table~\ref{tab2} and Table~\ref{tab3} present the anomaly detection and localization performance for each object in VisA and MVTec3D-RGB datasets, respectively.
{These two datasets contain difficult anomaly instances, such as candle, macaroni2, tire, and chewing-gum, that exhibit subtle variations in appearance compared to normal instances, posing significant challenges to anomaly detection.
Remarkably, our method achieves a 2.4\% I-AUC improvement in the ``candle'' category, a 4.6\% I-AUC improvement in the ``macarnoni2'' category, a 5.0\% I-AUC improvement in the ``tire'' category, and a 2.9\% PRO improvement in the chewing-gum category.} These results highlight the effectiveness of our approach in overcoming the limited discrimination capacity of pre-trained models and effectively identifying challenging anomalies.

\subsection{Ablation Study}
\textbf{Ablation of main components.}
In this part, we conduct an ablation study on the main components of our model: RAA module proposed in the first stage and HKD loss proposed in the second stage. In detail, RAA comprises a Matching-based Residual Gate (MRG) and Attribute-scaling Residual Generator (ARG). We utilize the RD~\cite{RD} as the baseline and compare the following settings: (a) directly utilizing the predicted anomaly weight from MRG as the anomaly score, (b) blindly generating residuals using ARG without considering the anomaly weight from MRG, (c) incorporating both MRG and ARG, \emph{i.e.}, the RAA module, (d) solely introducing the HKD loss to the baseline, and (e) employing all the components. {As shown in Table \ref{main_ablate} (a)-(d), the integration of MRG and ARG is essential for the advancement of the vanilla teacher model to achieve better performance. Solely relying on MRG leads to overfitting to synthetic anomalies, while replacing ARG with learnable residuals degrades performance. These will be further discussed in the subsequent ablation study (Tables~\ref{RAA utilize} and~\ref{ARG}).  Besides, without MRG (Exp. (c)), generating an equal proportion of residuals for both normal and abnormal samples disrupts the discrimination ability of the pre-trained model, resulting in an ambiguous representation space.}
The results of (d) and (f) show 
 the effectiveness of the hard knowledge distillation loss, which enhances the student's ability to accurately reconstruct the features of certain
fine-grained normal textures and rare normal patterns. By using all the components, our method can not only extend the pre-trained teacher's representation ability but also enhance the student's normal distillation ability, which satisfies the first and second assumptions in the distillation-based methods.

\textbf{Comparison of different alternative methods. }
Our framework augments the discrimination capacity of the pre-trained teacher model by leveraging synthetic anomalies. We compare our proposed method with alternative
discrimination-based approaches. {In experiments (1)-(4) of Table~\ref{RAA utilize}, we fix the proposed synthetic anomaly generation method and replace the RAA module with several alternative designs: (1) a conventional contrastive learning module to refine the pretrained features; (2) directly using MRG to predict the anomaly score; and (3) replacing MRG with a U-Net as a binary segmentation network. These alternatives all lead to performance degradation. The main reason is that these methods may overfit to the specific patterns of synthetic anomalies and fail to generalize to diverse real-world anomalies. In contrast, our RAA module avoids this pitfall through its adaptive residual learning mechanism, which suppresses residuals on normal samples to maintain the feature integrity of the pre-trained model.} In experiments (5)-(6), we use the anomaly synthesis method proposed by NSA, replace its encoder-decoder reconstruction framework with ours, and achieve improved performance, demonstrating the effectiveness and generalization of our method. In experiments (7)-(8), we integrate the memory bank of PatchCore into our AAND. Results show ``PatchCore+Ours" also outperforms PatchCore but underperforms our original version, as PatchCore's non-learnable nature prevents it from benefiting from synthetic anomaly learning.

\begin{table}[t]
\setlength\tabcolsep{2.0pt}
% \scriptsize
\centering
\caption{{Ablation Study of main modules of our framework. MRG and ARG are the sub-modules of RAA proposed in the first stage. $\mathcal{L}_{HKD}$ is the hard knowledge distillation loss proposed in the second stage.}}
\label{main_ablate}
\begin{tabular}{c|ccc|cc|cc|cc}
  \toprule
   \multirow{2}*{Exp.}&\multirow{2}*{MRG}& \multirow{2}*{ARG}& \multirow{2}*{$\mathcal{L}_{HKD}$}&  \multicolumn{2}{c|}{MvTec AD}&\multicolumn{2}{c|}{VisA}&\multicolumn{2}{c}{MvTec3D-RGB}\\
   % \cline{5-13}{0.1pt}
   % \Xcline{5-13}{0.01pt}
 % \cmidrule(r){5-7} \cmidrule(r){8-10} \cmidrule(r){11-13} 
 &&&& I-AUC&PRO& I-AUC&PRO& I-AUC&PRO\\
    \midrule
 \rowcolor{gray!12}
base&&&&98.5&94.0&96.2&93.4&88.5&96.7\\
\bf{(a)}&\checkmark& &&94.4&89.6&90.7&88.2&81.2&91.7\\
\bf{(b)}&\checkmark& $\checkmark\kern-1.2ex\raisebox{1ex}{\rotatebox[origin=c]{125}{\textbf{--}}}$&&98.2&94.8&94.5&94.2&85.6&96.0\\
\bf{(c)}&&\checkmark&&98.7&90.3&97.1&92.5&87.1&93.2\\
\bf{(d)}&\checkmark&\checkmark&&99.3&94.7&97.8&94.7&90.1&97.2\\
\bf{(e)}&&&\checkmark&99.1&94.3&97.4&93.7&89.6&96.6\\
\rowcolor{gray!12}
\bf{(f)}&\checkmark&\checkmark&\checkmark&\bf{99.5}&\bf{95.0}&\bf{98.1}&\bf{94.8}&\bf{90.6}&\bf{97.5}\\
\bottomrule
\end{tabular}
\end{table}

\begin{table}[t]
\setlength\tabcolsep{3.5pt}

% \scriptsize
\centering
\caption{{Comparison of Alternative Designs for the RAA Module.}}
\label{RAA utilize}
\begin{tabular}{l|cc|cc|cc}
  \toprule
 \multirow{2}*{Method}&  \multicolumn{2}{c|}{MvTec AD}&\multicolumn{2}{c|}{VisA}&\multicolumn{2}{c}{MvTec3D-RGB}\\
 % \cmidrule(r){3-5} \cmidrule(r){6-8}
& I-AUC&PRO& I-AUC&PRO& I-AUC&PRO\\
    \midrule
% (a) Binary classification &87.1&74.8&58.4&18.9&68.9&47.3\\
% (b) contrastive leaning&96.8&83.8&92.0&79.2&83.6&82.1\\写错了
(1) Syn.+Contrastive&73.7&43.3&61.2&23.4&59.9&34.2\\
(2) Syn.+Only MRG &94.4&89.6&90.7&88.2&81.2&91.7 \\
(3) Syn.+Unet+ARG&99.3& 94.4& 97.5& 93.5 & 88.8& 97.0\\
(4) Syn.+MRG+ARG&99.5&\bf{95.0}&\bf{98.1}&\bf{94.8}&\bf{90.6}&\bf{97.5}\\
\midrule
(5) NSA+Reconstruction&97.2&-&-&-&-&- \\
(6) NSA+Ours &\textbf{99.6}&94.7&97.7&94.3&89.0&97.1\\
\midrule
(7) PatchCore&99.1&94.4&95.1& 91.2&77.0&87.6\\
(8) PatchCore+Ours&99.4&94.2&97.2& 93.3&89.1&95.6\\
% (g) Ours \\
   \bottomrule
\end{tabular}
\end{table}

\begin{table}[t]
\setlength\tabcolsep{7.0pt}
\footnotesize
\centering
\caption{{Ablation of our two-stage training approach. The performances denote the I-AUC on the MvTec, VisA, and MVTec 3D-RGB datasets. }}
\label{3_2}
\begin{tabular}{c|c|c|c|c}
  \toprule
 Method&Framework&Performance&Infer. time&Param.   \\
 % \cmidrule(r){3-5} \cmidrule(r){6-8}
    \midrule
% (a) Binary classification &87.1&74.8&58.4&18.9&68.9&47.3\\
% (b) contrastive leaning&96.8&83.8&92.0&79.2&83.6&82.1\\写错了
RD&One Stage&98.5\,/\,\underline{96.2}\,/\,\underline{88.5}&\bf{0.024s}&\bf{161.1M}\\
RD++&One Stage&\underline{99.4}\,/\,96.0\,/\,88.2&0.027s&176.6M\\
\midrule
\multirow{4}*{Ours}&Stage I&94.4\,/\,90.7\,/\,81.2&0.016s&\,\,\,\,80.6M\\
% &Stage II&99.1\,/\,97.4\,/\,89.6&0.024s&161.1M\\
&Joint Train.&96.5\,/\,93.6\,/\,83.0&0.027s&172.8M\\
&Two Stage&\bf{99.5}\,/\,\bf{98.1}\,/\,\bf{90.6}&\underline{0.027s}&\underline{172.8M}\\

   \bottomrule
\end{tabular}
\end{table}

\textbf{Ablation of the two-stage framework.}
{
In this part, we conduct an ablation study on our two-stage framework. As shown in Table~\ref{3_2}, using Stage I alone leads to subpar performance as it cannot detect unforeseen anomalies, which requires Stage II to remember normality. Instead, integrating the two-stage training methodology fulfills the two underlying assumptions of the KD framework, culminating in optimal performance.
We also explore the possibility of joint training for the two stages, simultaneously optimizing both the advanced teacher and the student model. However, it yields poorer performances, which can be attributed to the continuous fluctuations in the teacher's feature space during training. These ongoing changes created significant challenges for the training of the student model, making it difficult to learn a stable normal feature distribution.
All in all, our two-stage training approach proves to be both reasonable and effective.}

\begin{figure}[h]
    \centering
\includegraphics[scale=0.35]{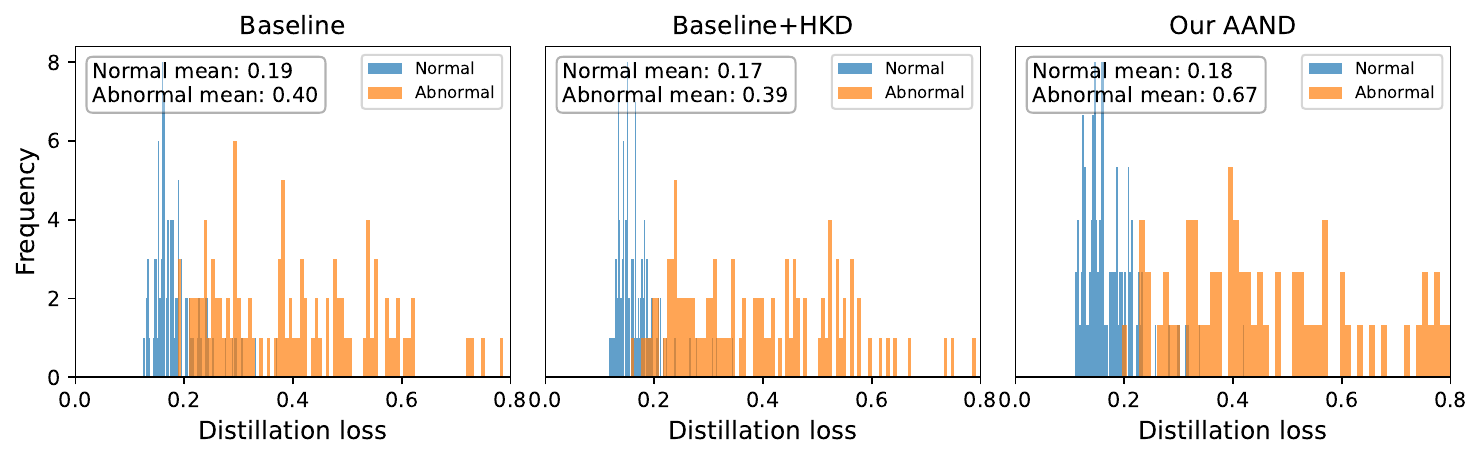}
    \caption{{Quantitative Experiments for the verification of the second assumption.}}
    \label{hist}
\end{figure}

{\textbf{Quantitative verification of Assumption II.} To verify the second assumption in our paper, we analyze the student’s distillation loss, as shown in Fig.~\ref{hist}. Adding the HKD loss to the baseline reduces the distillation loss of normal patches
by 0.02 (10.5\%), indicating that the student is better able to reconstruct normal features. With
our two-stage AAND method, normal and abnormal features are more clearly separated due to the fulfillment of the first assumption, making it harder for the student to reconstruct abnormal
features.}

% \begin{table}[t]
% \setlength\tabcolsep{4.0pt}
% \footnotesize
% \centering
% \caption{Comparison of distillation-based approaches in terms of inference time, network parameters, and performance (I-AUC/P-AUC/PRO) on the MVTec 3D-RGB dataset.}
% \label{time}
% \begin{tabular}{l|c|c|c}
%    \toprule
%   Method&Infer. time (s)& Parameters (MB)&Performance (\%) \\
%   \midrule
%  RD&\bf{0.024}&\bf{161.08}&88.5~/~99.0~/~96.7\\
%   RD++ &0.027&176.57&88.2~/~99.0~/~97.2\\
  
% Ours&0.027&170.45&\bf{90.6}~/~\bf{99.2}~/~\bf{97.5}\\
%    \bottomrule
% \end{tabular}
% \end{table}

\begin{table}[t]
\setlength\tabcolsep{3.5pt}
% \scriptsize
\centering
\caption{Analysis of the residual properties in the ARG module.}
\label{ARG}
\begin{tabular}{l|cc|cc|cc}
  \toprule
 \multirow{2}*{Method}&  \multicolumn{2}{c|}{MvTec AD}&\multicolumn{2}{c|}{VisA}&\multicolumn{2}{c}{MvTec3D-RGB}\\
 % \cmidrule(r){3-5} \cmidrule(r){6-8}
& I-AUC&PRO& I-AUC&PRO& I-AUC&PRO\\
    \midrule
(a) Learnable residuals &98.2&94.8&94.5&94.2&85.6&96.0\\
% (b) ARG w/o $\mathbf{S}^{ref}$&99.4&94.9&97.9&94.7&90.3&97.4\\
(b) ARG w/ sigmoid &99.5&94.4&97.6&94.2&90.0&97.0\\
(c) ARG w/ tanh &\bf{99.5}&\bf{95.0}&\bf{98.1}&\bf{94.8}&\bf{90.6}&\bf{97.5}\\
   \bottomrule
\end{tabular}
\end{table}

\begin{figure*}[th]
    \centering
\includegraphics[scale=0.68]{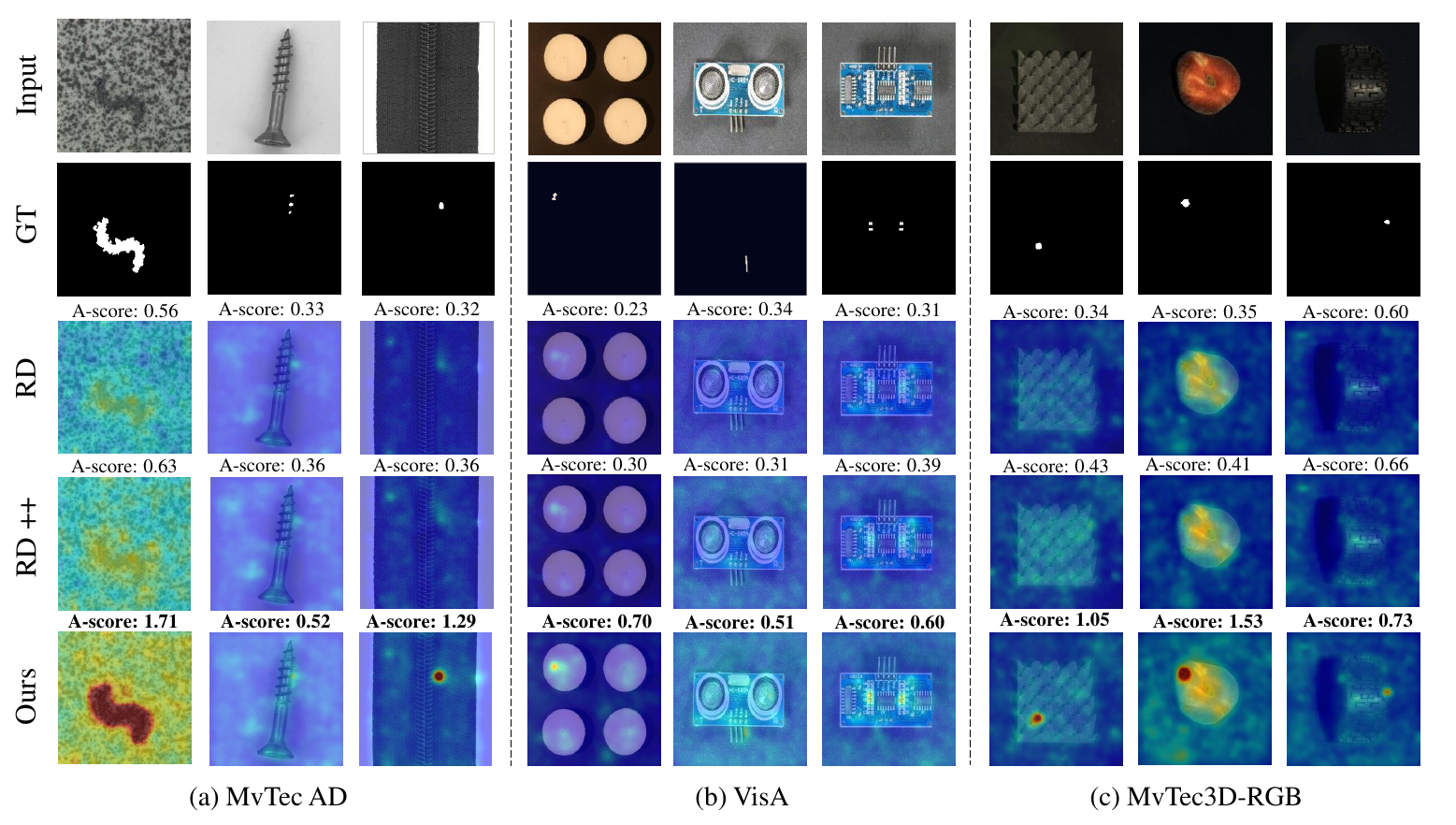}
    \caption{Qualitative results for anomaly localization, where ``A-score" denotes the maximum value in the anomaly score map.
    Compared to RD~\cite{RD} and RD++~\cite{RDplus}, our method can accurately localize anomalies even in some challenging cases where the abnormal region is extremely small or the appearance of the anomaly is very similar to normal data.}
    \label{vis_main}
\end{figure*}

\begin{figure*}[t]
	\centering
	\begin{minipage}{0.45\linewidth}
		% \vspace{4pt}
\centerline{\includegraphics[width=\textwidth]{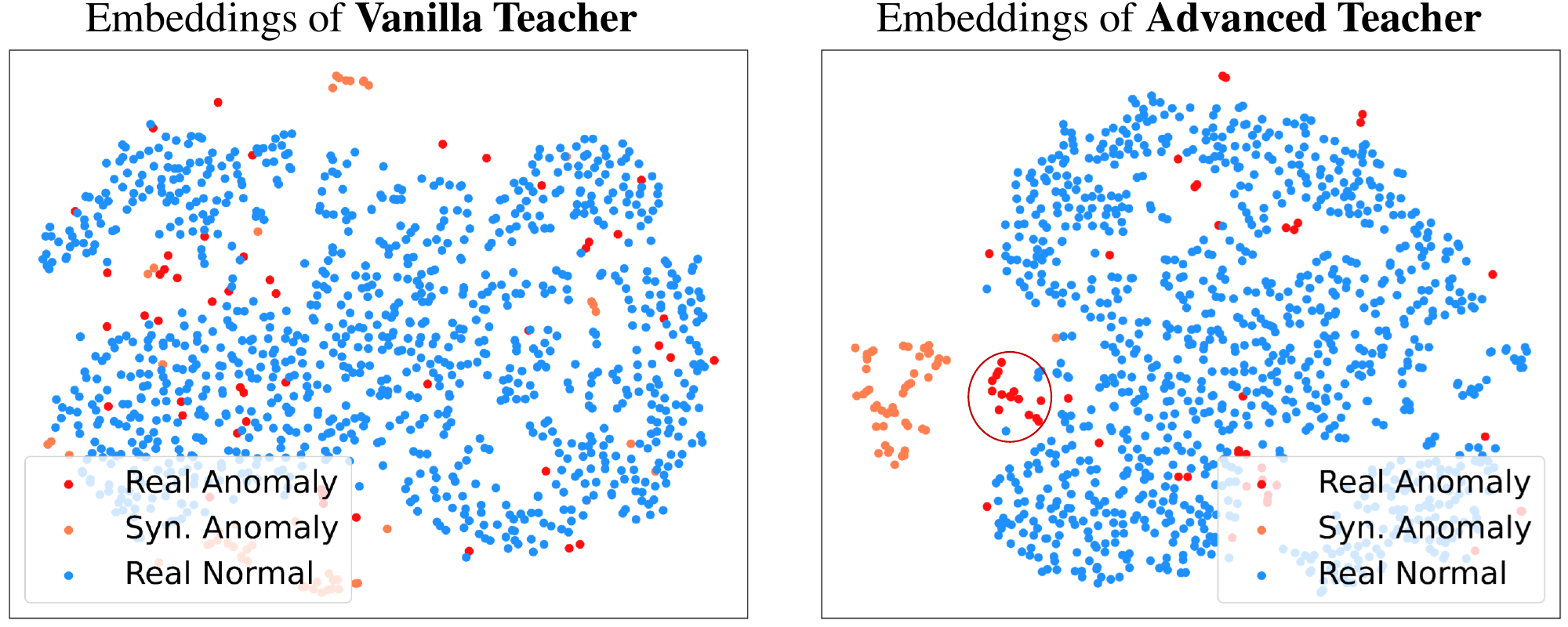}}
          % 加入对这列的图片说明
	 % \vspace{-0.3cm}
     
     \centerline{\footnotesize{(a) candle class} }
     \vspace{0.3cm}
	\end{minipage}\quad
	\begin{minipage}{0.45\linewidth}
		% \vspace{4pt}
\centerline{\includegraphics[width=\textwidth]{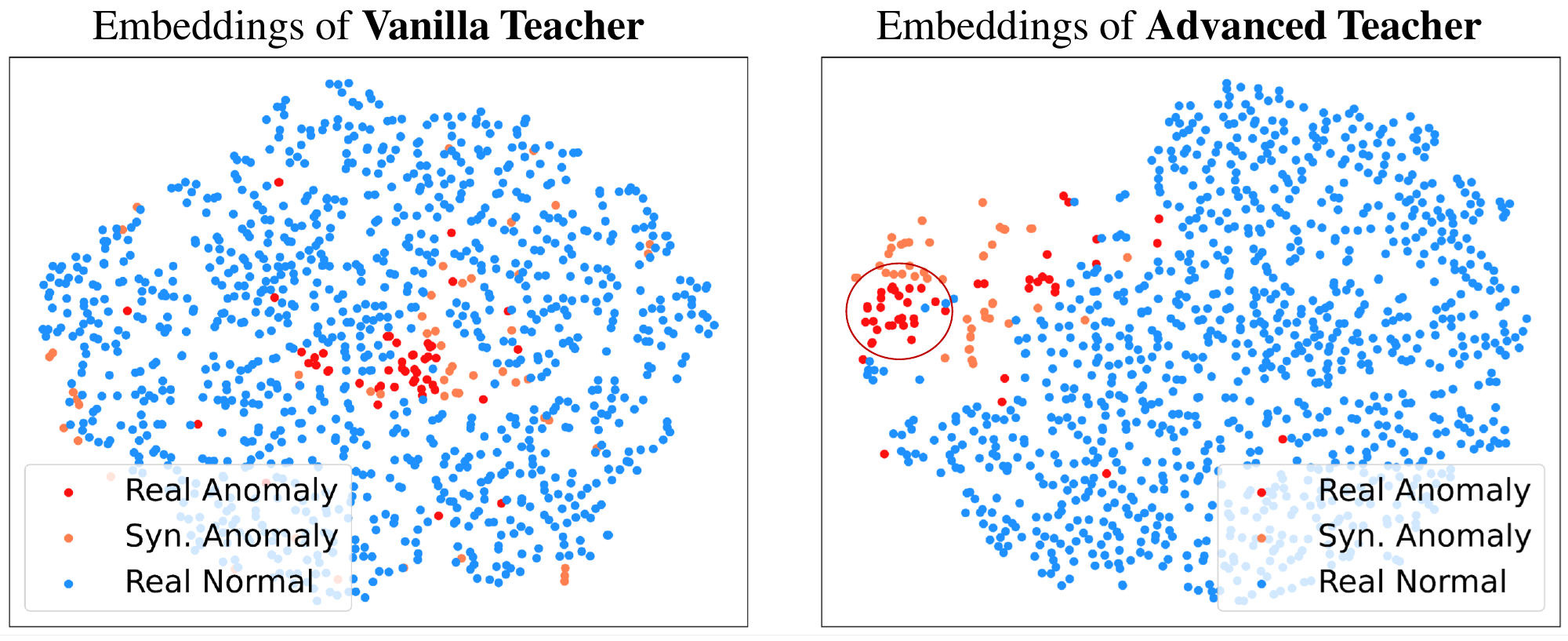}}
  % \vspace{-0.3cm}
\centerline{\footnotesize{(b) macaroni2 class}}
\vspace{0.3cm}
	\end{minipage}

% \vspace{0.4cm}
 \centering
	\begin{minipage}{0.45\linewidth}
		% \vspace{4pt}
\centerline{\includegraphics[width=\textwidth]{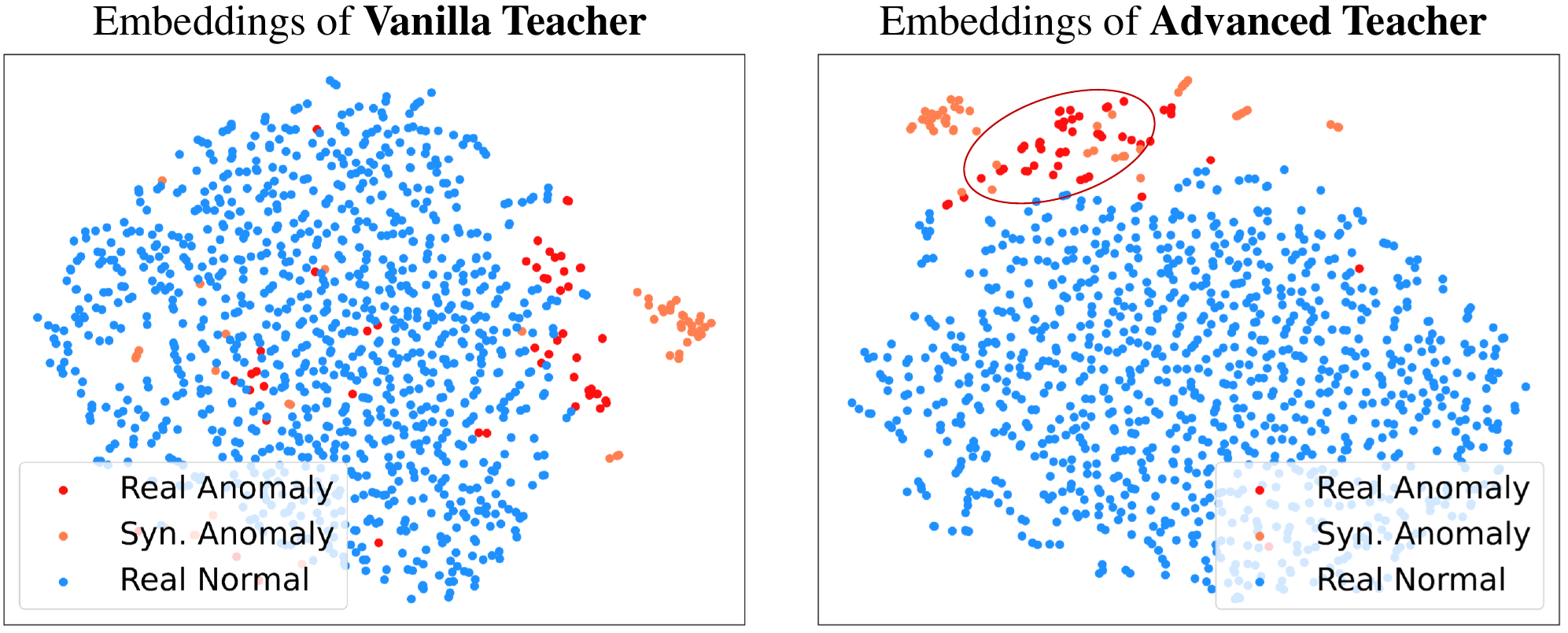}}
          % 加入对这列的图片说明
% \vspace{-0.3cm}
\centerline{\footnotesize{(c) bagel class} }
\vspace{0.3cm}
	\end{minipage}\quad
	\begin{minipage}{0.45\linewidth}
		% \vspace{4pt}
\centerline{\includegraphics[width=\textwidth]{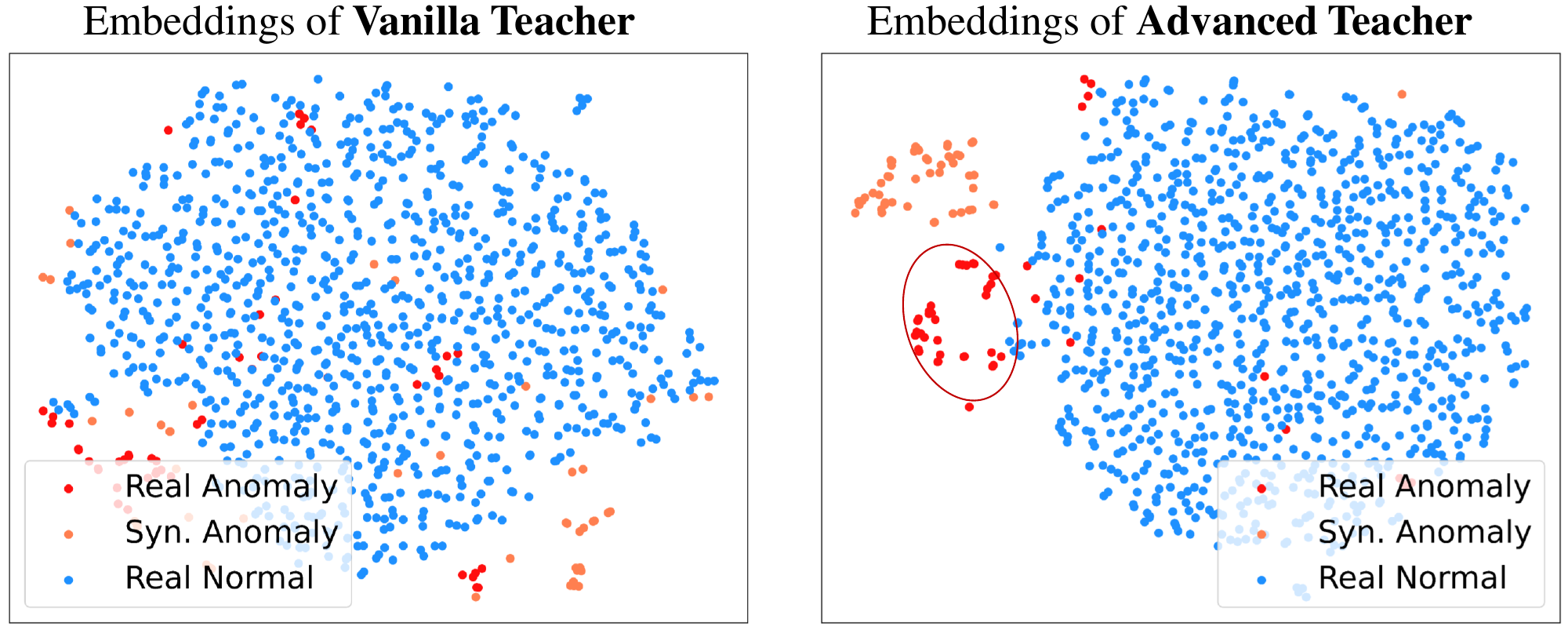}}
  % \vspace{-0.3cm}
\centerline{\footnotesize{(d) foam class}}
\vspace{0.3cm}
	\end{minipage}

    % \vspace{0.4cm}
    \begin{minipage}{0.45\linewidth}
		% \vspace{4pt}
\centerline{\includegraphics[width=\textwidth]{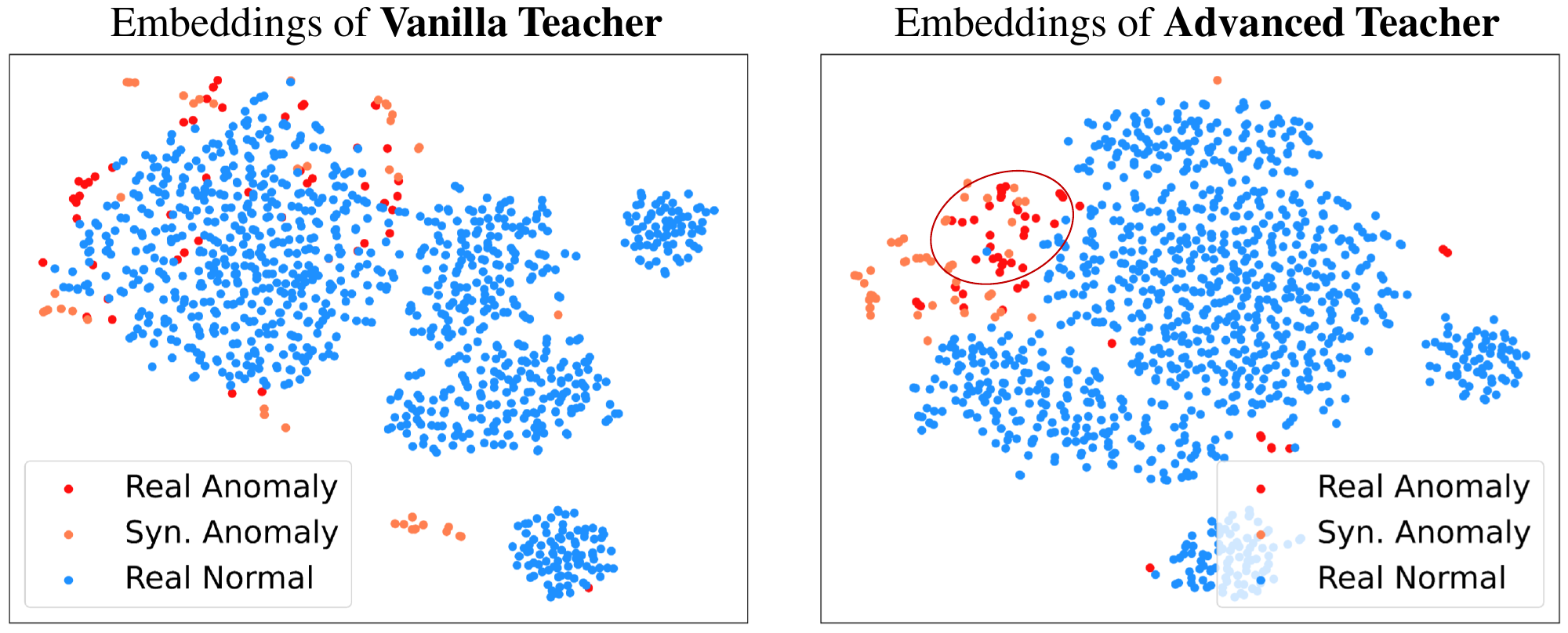}}
% \vspace{-0.3cm}
\centerline{\footnotesize{(e) carpet class}}
\vspace{0.3cm}
	\end{minipage}\quad
    \begin{minipage}{0.45\linewidth}
		% \vspace{4pt}
\centerline{\includegraphics[width=\textwidth]{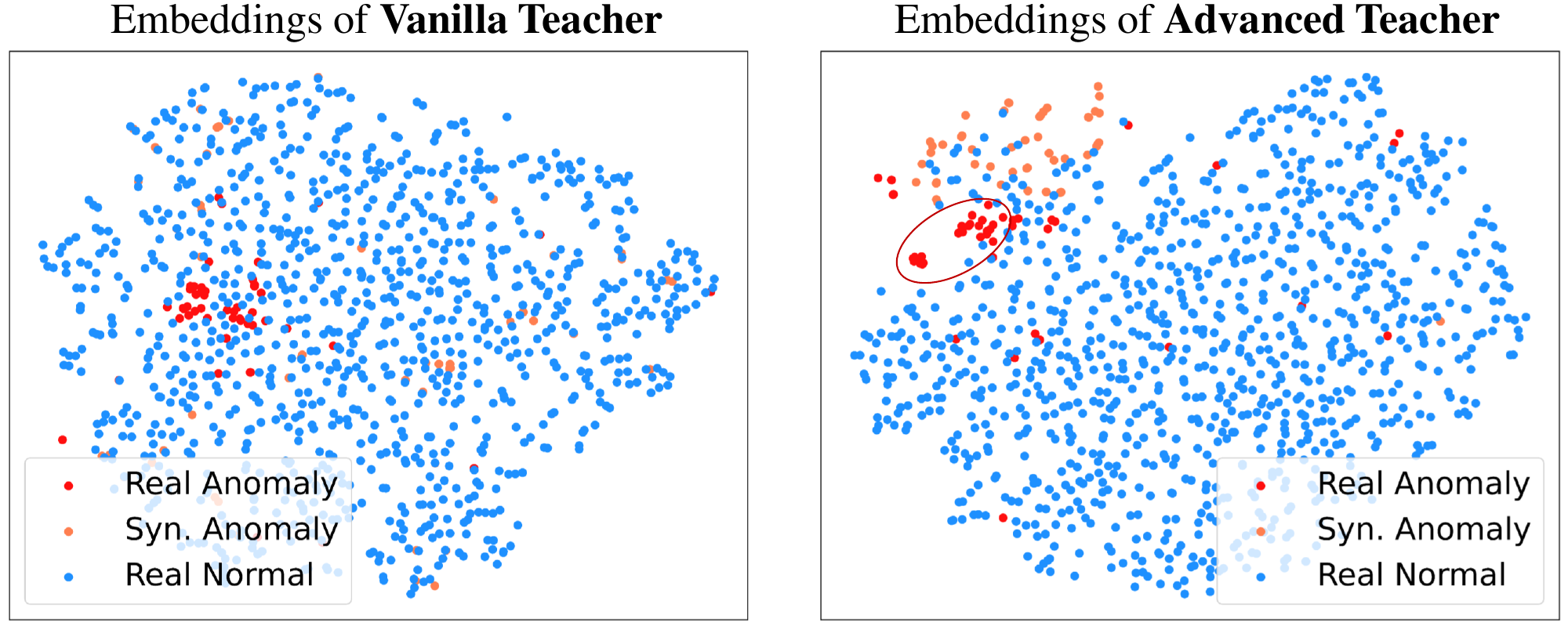}}
% \vspace{-0.3cm}
\centerline{\footnotesize{(f) pcb1 class}}
\vspace{0.3cm}
	\end{minipage}
	\caption{{Comparison of feature distribution between the vanilla teacher model and our advanced teacher model,  visualized using t-SNE~\cite{tsne}.
Figure (a)-(f) represent specific industrial product categories from the benchmark datasets. Our method demonstrates an enhanced discrimination capacity between real normal and abnormal inputs while preserving the integrity of pre-trained normal representation. The red circle marks out the optimized anomaly features.}}
	\label{tsne_RAA}
\end{figure*}

\begin{table}[t]
\setlength\tabcolsep{10pt}
\footnotesize
\centering
\caption{Performance comparison of CNN-based and Transformer-based frameworks. The performances denote the I-AUC on the MvTec, VisA, and MvTec3D-RGB datasets.}
\label{backbone}
\begin{tabular}{ccc}
  \toprule
 Method&Pre-trained Model& Performance\\
 % \cmidrule(r){3-5} \cmidrule(r){6-8}
    \midrule
% (a) Binary classification &87.1&74.8&58.4&18.9&68.9&47.3\\
% (b) contrastive leaning&96.8&83.8&92.0&79.2&83.6&82.1\\写错了
RD\,(CNN)&Wide-Resnet50-2\,(69\,M)&98.5\,/\,96.2\,/\,88.5\\
Ours\,(CNN)&Wide-Resnet50-2\,(69\,M)&\bf{99.5}\,/\,\bf{98.1}\,/\,\bf{90.6}\\
\midrule
RD\,(Tran.)&CLIP-ViT-B/16\,(149\,M)&95.2\,/\,90.1\,/\,81.0\\
Ours\,(Tran.)&CLIP-ViT-B/16\,(149\,M)&97.2\,/\,92.5\,/\,82.6\\

   \bottomrule
\end{tabular}
\end{table}

\textbf{Analysis of residual properties. }
To understand the impact of different residual characteristics in the ARG module, we conduct ablation studies that are shown in Table~\ref{ARG}. It involves (a) replacing the residuals in Eq.~\eqref{delta} with learnable residuals, and (b) replacing the tanh function in Eq.~\eqref{delta} with a sigmoid function.
The utilization of learnable residuals without explicit scale restrictions can potentially result in excessive residuals and the disruption of the pre-trained model's prior knowledge. In contrast to scaling the residuals to the range (0, 1) using the sigmoid function, we employ the tanh function to project the residuals to the range of (-1, 1). This choice leads to moderate perturbations to the original features, thus improving the performance effectively.

\textbf{Ablation of backbone.}
{To our knowledge, no prior work has explored transformer-based architectures within the reverse distillation~\cite{RD} framework.  As a pilot exploration, we replace the original pre-trained
Wide-Resnet50-2 encoder with a pre-trained CLIP-ViT-B/16 encoder and transform the
decoder from convolutions to ViT blocks. Under the same training setting,
the experimental results are shown in Tabel~\ref{backbone}. Whether using CNN-based or Transformer-based
architectures, our method achieves significant improvements over the Baseline RD, demonstrating
the effectiveness of our approach. However, the performance of the Transformer-based architecture currently lags behind that of the CNN-based method within the KD-based framework. This may be because the downsample-upsample structure of the CNN-based RD framework aligns more closely with the concept of reverse distillation. Addressing
this issue may require a specialized design for the transformer decoder, which could be explored in future work.}

 \begin{table}[t]
\setlength\tabcolsep{5.5pt}
\footnotesize
\centering
\caption{{Comparative analysis of model efficiency (FlOPs) and computational performance (MvTec / VisA / MvTec3D-RGB).}}
\label{flops}
\begin{tabular}{cccccc}
   \toprule
 Method & Teacher & \textbf{Additional} & Student (+BN) & Performance  \\
  \midrule
  RD&\textbf{12.03G}&\textbf{0}&\textbf{18.50G}&98.5\,/\,96.2\,/\,88.5\\
 RD++&12.03G&9.08G&18.50G&99.4\,/\,96.0\,/\,88.2\\
 Ours*&\underline{12.03G}&\underline{5.55G}&\underline{18.50G}&\underline{99.4}\,/\,\underline{97.8}\,/\,\underline{90.3}\\
 Ours&12.03G&10.38G&18.50G&\bf{99.5}\,/\,\bf{98.1}\,/\,\bf{90.6}\\
   \bottomrule
\end{tabular}
\end{table}

\textbf{Model complexity analysis. } 
 We compare the inference speed and number of parameters between our approach and existing distillation-based methods. The experiment is conducted on a Nvidia GeForce GTX 2080ti GPU with Intel(R)
Xeon(R) CPU E5-2680 v4@ 2.40GHZ and the batch size is set to 1 during testing.  As shown in Table~\ref{3_2}, our method introduces a marginal increase in inference time compared to RD and possesses fewer parameters than RD++. In comparison to the one-stage baseline RD, our two-stage framework solely incorporates the lightweight RAA modules, resulting in a slight computational overhead but a remarkable enhancement in performance. {We further evaluate computational efficiency in terms of FLOPs, as shown in Table~\ref{flops}. Compared to RD++, our method introduces 1.3 GFLOPs overhead but achieves performance improvements on challenging benchmarks (\textit{VisA: +2.1\%, MvTec3D-RGB: +2.4\%}). Additionally, we offer a lightweight variant Ours* that uses 2-layer MLPs instead of 3-layer MLPs in the RAA module, reducing computational cost by 4.83 GFLOPs while retaining gains (\textit{VisA: +1.8\%, MvTec3D-RGB: +2.1\%}). These results demonstrate that our approach achieves a favorable efficiency-accuracy trade-off, obtaining superior performance with lower computational cost.}

% The learnable residuals have no explicit scale restrictions, potentially leading to excessive residuals and disruption of the pre-trained model's prior knowledge.
% The sigmoid function limits the residual scaling factor to the range (0, 1) instead of (-1,1), resulting in smaller residual scaling factors and less significant enhancement of the pre-trained features.
% (-1,1) 

\begin{figure}[th]
    \centering
    % \hspace{-0.5cm}
    \includegraphics[scale=0.18]{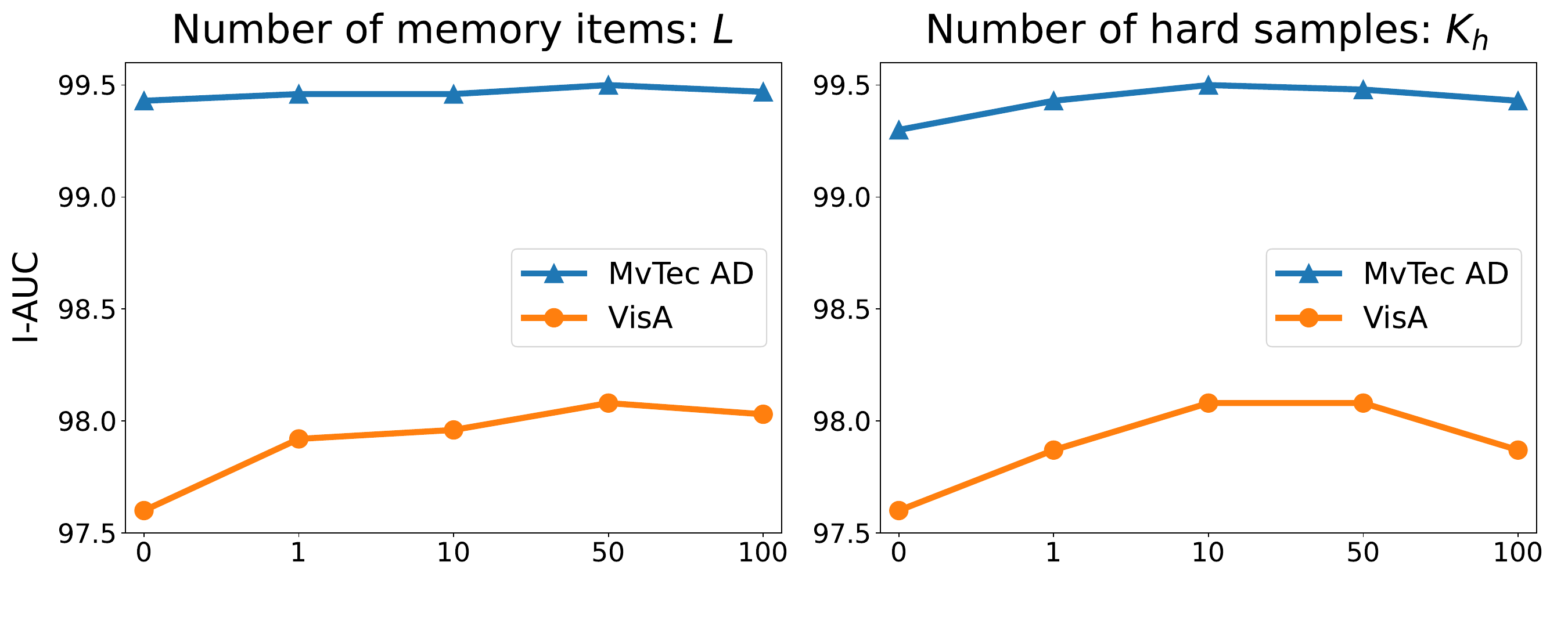}
    \caption{Hyper-parameter sensitivity analysis for the number of memory items $L$ and the number of hard samples $K_h$.}
    \label{sensi}
\end{figure}

\textbf{Hyper-parameter sensitivity.}
There are two important hyper-parameters in our method: the number of memory items $L$ in MRG module, and the number of the selected hard sample $K_h$ in the hard knowledge distillation loss. We evaluate the sensitivity of our model to the two parameters on the MvTec AD~\cite{mvtec} and VisA~\cite{visa} datasets. We first fix $K_h$ at 10 to change $L$, and then fix $L$ to the best value to change $K_h$, where $L=0$ means that we use MLP as a classifier instead of MRG. As shown in Fig.~\ref{sensi}, MRG demonstrates robustness concerning the memory size $L$. Appropriately increasing the memory size helps to capture the diversity of normality and anomaly, where the best performance is achieved when the memory size is set to 50.
For the number of hard samples $K_h$, compared to not selecting hard samples (\emph{i.e.}, $K_h=0$), selecting different numbers of hard samples can improve performance. The best performance is achieved when $K_h=10$. Insufficient hard samples lack representativeness of challenging patterns, while an excessive amount of hard samples hinders the model's focus on the most challenging patterns.

%   \begin{figure}[th]
%     \centering
%     % \hspace{1.0cm}
%     \includegraphics[scale=0.39]{vis2.pdf}
%     \caption{The histogram of predicted anomaly scores for normal samples and abnormal samples during inference for MVTec3D-RGB dataset (x-axis: anomaly score, and y-axis: count).}
%     \label{fig4}
% \end{figure}
% \textbf{Synthesis-based discrimination. }
% \textbf{Single-stage distillation. }
% \textbf{Two-stage distillation. }

\begin{figure}[th]
    \centering
    \includegraphics[scale=0.18]{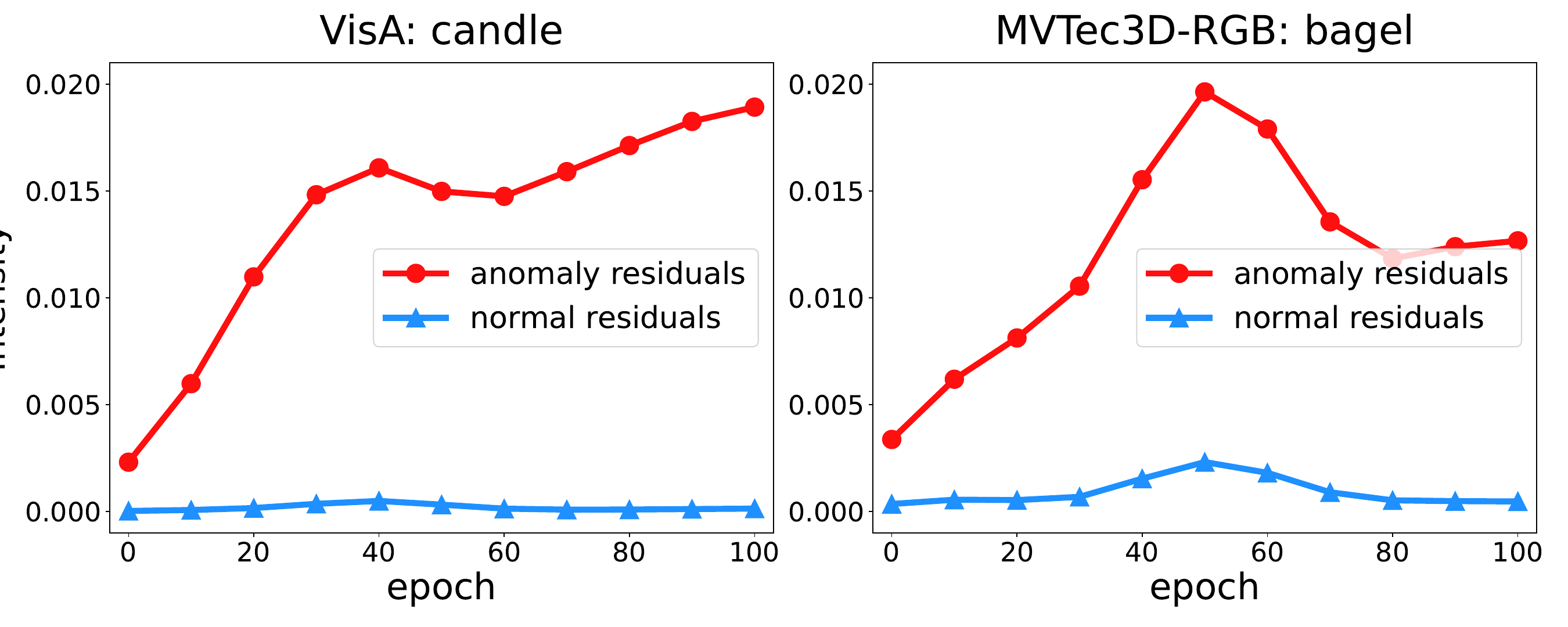}
    \caption{The variation of the intensity of residuals with respect to training
epochs. The blue line indicates the residual variation on normal samples, while the red line indicates the residual variation on anomaly samples.}
    \label{res_vis}
\end{figure}

\subsection{Visualization Analysis}
\textbf{Qualitative results for anomaly localization. }
We present qualitative anomaly localization results in Fig.~\ref{vis_main}. Compared to RD and RD++, our method can accurately localize anomalies even in some challenging cases, where the abnormal region is extremely small or the appearance of the anomaly is very similar to normal data.
When dealing with these cases, the vanilla teacher model in RD and RD++ maps abnormal inputs to a normal representation, in which the student also constructs the normality. As a result, some defective products cannot be successfully detected due to the unreliable teacher-student feature discrepancy.
Unlike these methods, our advanced teacher model can generate more discriminative features, which better fulfills the first assumption in distillation-based anomaly detection methods. In practice, detecting these challenging anomalies helps manufacturers identify problems promptly and ensures that consumers receive satisfactory products.
% However, in the third and fourth rows of images, some specific anomaly patches are still not detected for the compared methods and ours. This is because the geometric variations in these cases are not visible in RGB images. This could be addressed by incorporating point cloud or depth information as ~\cite{M3DM, AST}, yet it introduces additional computation cost.

% 
\textbf{Visualization of feature distribution.}
To demonstrate the enhanced discrimination capacity of our anomaly amplification stage, we present the visualization of feature distribution in Fig.~\ref{tsne_RAA}.  {Without the introduced stage, the feature distribution of the vanilla teacher exhibits ambiguity between normal and abnormal features. In our advanced teacher space, the feature distribution is significantly more separated than that in the vanilla teacher, which proves the effectiveness of our Stage I. Moreover, real anomalies can be well distinguished from normal ones, demonstrating that synthetic anomalies can approximate a majority of real anomalies.
Additionally, the shape of the normal distribution does not undergo significant changes before and after Stage I training. This is because our residual gate suppresses residuals on normal samples, preserving the generalizability towards diverse normal patterns.}

\textbf{Visualization of residual variation.}
Our proposed RAA module suppresses the residuals on normal features while encouraging residuals for abnormal features. To demonstrate this, we present visualizations of residual variations during training in Fig.~\ref{res_vis}. We define the intensity of residual noises for normal samples and anomaly samples as $\frac{1}{N^n\times C}\sum_{i=1}^{N^n}\sum_{j=1}^{C}{(\Delta^n_{ij})^2}$ and $\frac{1}{N^a\times C}\sum_{i=1}^{N^a}\sum_{j=1}^{C}{(\Delta^a_{ij})^2}$, respectively.
During the training process, the intensity of residual noise is kept minimal for normal samples, thereby preserving the pre-trained model's fidelity to normal instances. Conversely, the residuals on abnormal inputs exhibit an increasing trend, amplifying the distinctions between anomalies and normal instances.
% \textbf{visualization of dual memory. }
% As shown in \ref{fig7}, we provide a visualization of the normal memory and the anomaly memory at different stages of the feature extraction process. In general, they show multimodal distributions within classes, while exhibiting distinguishable patterns between classes. From level 1 to level 3, the differentiation between the normal memory and the anomaly memory becomes more significant. This divergence arises due to the increasing representation ability of the query, allowing for obvious discrimination between normal and abnormal patterns.

% \begin{figure}[th]
%     \centering
%     % \hspace{-0.5cm}
%     \includegraphics[scale=0.18]{RAR_memory.pdf}
%     \caption{T-SNE \cite{tsne} visualization of normal memory and anomaly memory at different stages of the feature extraction process. The first row to the third row shows the results of bagel, cable gland, and carrot objects from the MVTec3D-RGB dataset \cite{mvtec3d}.}
%     \label{fig7}
% \end{figure}

\section{Conclusion}\label{conclu}
In this paper, we propose a novel two-stage industrial anomaly detection
framework, which sequentially performs
anomaly amplification and normality distillation to obtain
robust feature discrepancy. 
In the first anomaly amplification
stage, we propose a novel residual anomaly amplification module to advance the teacher model with the exposure of synthetic anomalies. It effectively amplifies anomalies via residual generation while maintaining the integrity of the pre-trained model. In the second normality distillation stage, we design a novel hard knowledge distillation loss, which enhances the
reconstruction capability of the student decoder in dealing
with challenging normal patterns. Comprehensive experiments on the MvTecAD, VisA, and
MvTec3D-RGB dataset demonstrate the effectiveness of our
method. 

\textbf{Limitation and future work:} Some hard anomalies are still difficult to detect solely from RGB images. Combining knowledge from other modalities, such as point clouds and natural language, may lead to better detection results. Besides, synthesizing anomaly data that closely resembles real-world anomalies is crucial for better advancing pre-trained models. {In addition, decomposing pretrained teacher model into task-specific factor networks~\cite{yang2022factorizing} is also a promising direction.}
These will be explored in our future work.

% \begin{thebibliography}{1}
\bibliographystyle{IEEEtran}
\bibliography{ref}

\vfill

\end{document}